%% file: main.tex
\documentclass[10pt,journal,compsoc]{IEEEtran}

\ifCLASSOPTIONcompsoc
  \usepackage[nocompress]{cite}
\else
  \usepackage{cite}
\fi

\ifCLASSINFOpdf
   \usepackage[pdftex]{graphicx}
   \DeclareGraphicsExtensions{.pdf,.jpeg,.png}
\else
\fi

\usepackage{times}
\usepackage{epsfig}
\usepackage{graphicx}
\usepackage{amsmath}
\usepackage{amssymb}
\usepackage{booktabs}
\usepackage{color, colortbl}
\usepackage[dvipsnames]{xcolor}
\usepackage{multirow}

\usepackage{tikz}
\usepackage{pgfplots}
\usepackage{pgfplotstable}
\usepackage{subcaption}
\usepackage{mathtools}

\usepackage[colorlinks,citecolor=red,urlcolor=blue,bookmarks=false,hypertexnames=true]{hyperref}

\usepackage{array}
\newcolumntype{P}[1]{>{\centering\arraybackslash}p{#1}}
\newcolumntype{M}[1]{>{\centering\arraybackslash}m{#1}}

\newcommand{\figref}[1]{Fig.~\ref{#1}}
\newcommand{\secref}[1]{Sec.~\ref{#1}}
\newcommand{\tabref}[1]{Table~\ref{#1}}

\makeatletter
\DeclareRobustCommand\onedot{\futurelet\@let@token\@onedot}
\def\@onedot{\ifx\@let@token.\else.\null\fi\xspace}
\def\eg{\emph{e.g}.} 
\def\ie{\emph{i.e}.}

\makeatother

\begin{document}

% \title{Bare Demo of IEEEtran.cls for\\ IEEE Computer Society Journals}

\title{Transformers in Action: \\ Weakly Supervised Action Segmentation}

\newcommand{\methodname}{TransAcT}
\definecolor{themered}{HTML}{FF595E}
\definecolor{themeorange}{HTML}{FF8E3F}
\definecolor{themeyellow}{HTML}{FFC31F}
\definecolor{themegreen}{HTML}{76AB21}
\definecolor{themeblue}{HTML}{1982C4}
\definecolor{themeindigo}{HTML}{4267AC}
\definecolor{themepurple}{HTML}{6A4C93}

\author{John~Ridley, 
        Huseyin~Coskun,
        David Joseph Tan,
        Nassir~Navab,~\IEEEmembership{Fellow,~IEEE},
        Federico~Tombari% <-this % stops a space
\IEEEcompsocitemizethanks{
\IEEEcompsocthanksitem John Ridley is at Technische Universit\"at M\"unchen.\protect\\
E-mail: john.ridley@tum.de

\IEEEcompsocthanksitem H. Coskun is at Technische Universit\"at M\"unchen.\protect\\
E-mail: huseyin.coskun@tum.de
\IEEEcompsocthanksitem DJ. Tan is at Google.
\IEEEcompsocthanksitem  N. Navab is at Technische Universit\"at M\"unchen. % <-this % stops an unwanted space
\IEEEcompsocthanksitem F. Tombari is at Google and Technische Universit\"at M\"unchen.

}% <-this % 
}

\IEEEtitleabstractindextext{%
\begin{abstract}
  The video action segmentation task is regularly explored under weaker forms of supervision, such as transcript supervision, where a list of actions is easier to obtain than dense frame-wise labels. In this formulation, the task presents various challenges for sequence modeling approaches due to the emphasis on action transition points, long sequence lengths, and frame contextualization, making the task well-posed for transformers. Given developments enabling transformers to scale linearly, we demonstrate through our architecture how they can be applied to improve action alignment accuracy over the equivalent RNN-based models with the attention mechanism focusing around salient action transition regions. Additionally, given the recent focus on inference-time transcript selection, we propose a supplemental transcript embedding approach to select transcripts more quickly at inference-time. Furthermore, we subsequently demonstrate how this approach can also improve the overall segmentation performance. Finally, we evaluate our proposed methods across the benchmark datasets to better understand the applicability of transformers and the importance of transcript selection on this video-driven weakly-supervised task.
\end{abstract}

}

\maketitle

\IEEEdisplaynontitleabstractindextext
\IEEEpeerreviewmaketitle

\input{sections/1-intro.tex}
\input{sections/2-related.tex}
\input{sections/3-method}
\input{sections/4-experiment.tex}
\input{sections/5-conclusion.tex}
% % use section* for acknowledgment
% \ifCLASSOPTIONcompsoc
%   % The Computer Society usually uses the plural form
%   \section*{Acknowledgments}
% \else
%   % regular IEEE prefers the singular form
% \fi

% The authors would like to thank xxx for the valuable discussions and constructive feedback. \cHu{We should delete this section if you don't have anyone or any company to thank}

\ifCLASSOPTIONcaptionsoff
  \newpage
\fi



%\bibliographystyle{IEEEtran}

{\small
\bibliographystyle{IEEEtran}
\bibliography{bibliography}
}
\vspace{-5mm}

\begin{IEEEbiography}[{\includegraphics[width=1.1in,height=1.5in,clip,keepaspectratio]{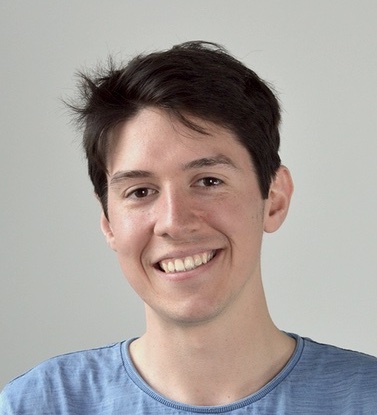}}]{John Ridley} is a Researcher at Technische Universit\"at M\"unchen. He obtained his B. Sc. degree in Computer Science with Honours from Curtin University in 2017, and his M.Sc. degree in Informatics from Technische Universit\"at M\"unchen, with a focus on deep learning and metric learning applied to computer vision tasks.
\end{IEEEbiography}

\begin{IEEEbiography}[{\includegraphics[width=1.1in,height=1.5in,clip,keepaspectratio]{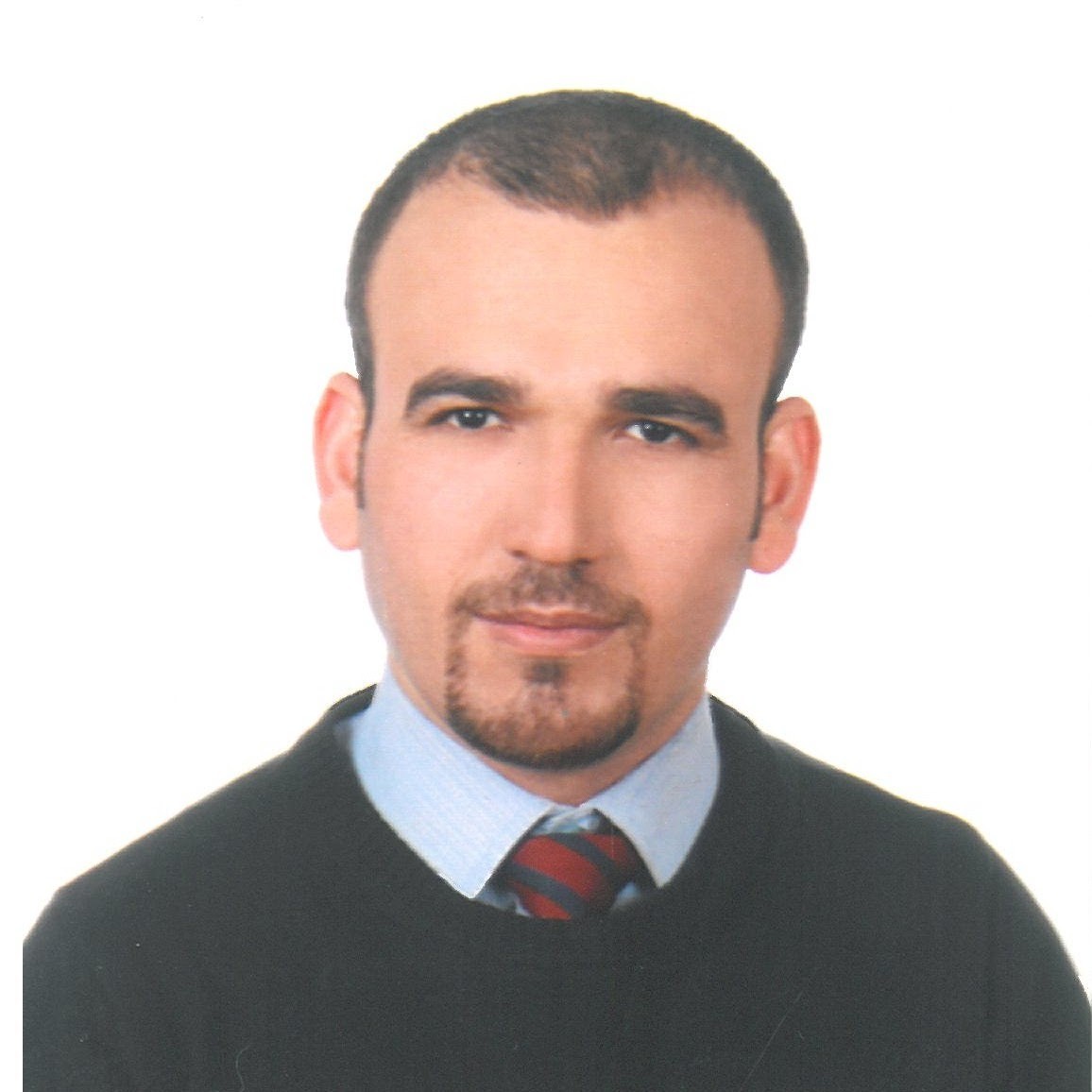}}]{Huseyin Coskun} is a PhD candidate at the Computer Aided Medical Procedures and Augmented Reality (CAMP) laboratories at Technische Universit\"at M\"unchen. He obtained his B. Sc. degree in Mathematics from Instanbul Technical University in 2011, and his M.Sc. degree in Artificial Intelligence from Polytechnic University of Catalonia. He interned two times in Hololens/Vision at Microsoft. His work focuses on activity recognition, few-shot learning and meta learning.
\end{IEEEbiography}

\begin{IEEEbiography}[{\includegraphics[width=1.1in,height=1.5in,clip,keepaspectratio]{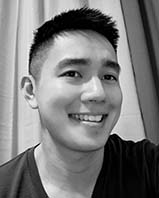}}]{David Joseph Tan}
is a Research Scientist at Google. He got his Ph.D.~from the Technische Universit\"{a}t M\"{u}nchen (TUM). In 2017, the culmination of his Ph.D.~research on the ultra-fast 6D pose estimation garnered him the Best Demo Award at ISMAR, EXIST-Gr\"{u}nderstipendium from the German government and the Promotionspreise from TUM.
Expanding his research in computer vision and machine learning, he continuously publishes in top-tier conferences and journals.
\end{IEEEbiography}

\begin{IEEEbiography}[{\includegraphics[width=1.1in,height=1.5in,clip,keepaspectratio]{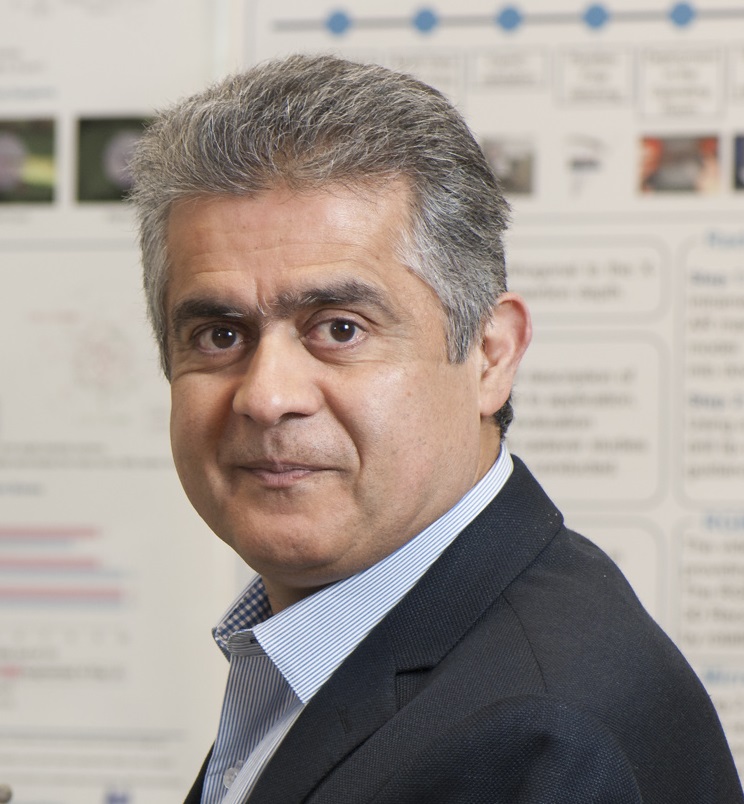}}]{Nassir Navab} is a full Professor and Director of the Laboratory for Computer Aided Medical Procedures, Technical University of Munich and Johns Hopkins University. He has also secondary faculty appointments at both affiliated Medical Schools. He completed his PhD at INRIA and University of Paris XI, France, and enjoyed two years of post-doctoral fellowship at MIT Media Laboratory before joining Siemens Corporate Research (SCR) in 1994. At SCR, he was a distinguished member and received the Siemens Inventor of the Year Award in 2001. He received the SMIT Society Technology award in 2010 and the ‘10 years lasting impact award’ of IEEE ISMAR in 2015. In 2012, he was elected as a Fellow of the MICCAI Society. He has acted as a member of the board of directors of the MICCAI Society and serves on the Steering committee of the IEEE Symposium on Mixed and Augmented Reality (ISMAR) and Information Processing in Computer Assisted Interventions (IPCAI). He is the author of hundreds of peer reviewed scientific papers, with more than 38700 citations and an h-index of 89 as of January 2020. He is the inventor of 50 granted US patents and more than 50 International ones. His current research interests include medical augmented reality, computer-aided surgery, medical robotics, and machine learning.
\end{IEEEbiography}

\begin{IEEEbiography}[{\includegraphics[width=1.1in,height=1.5in,clip,keepaspectratio]{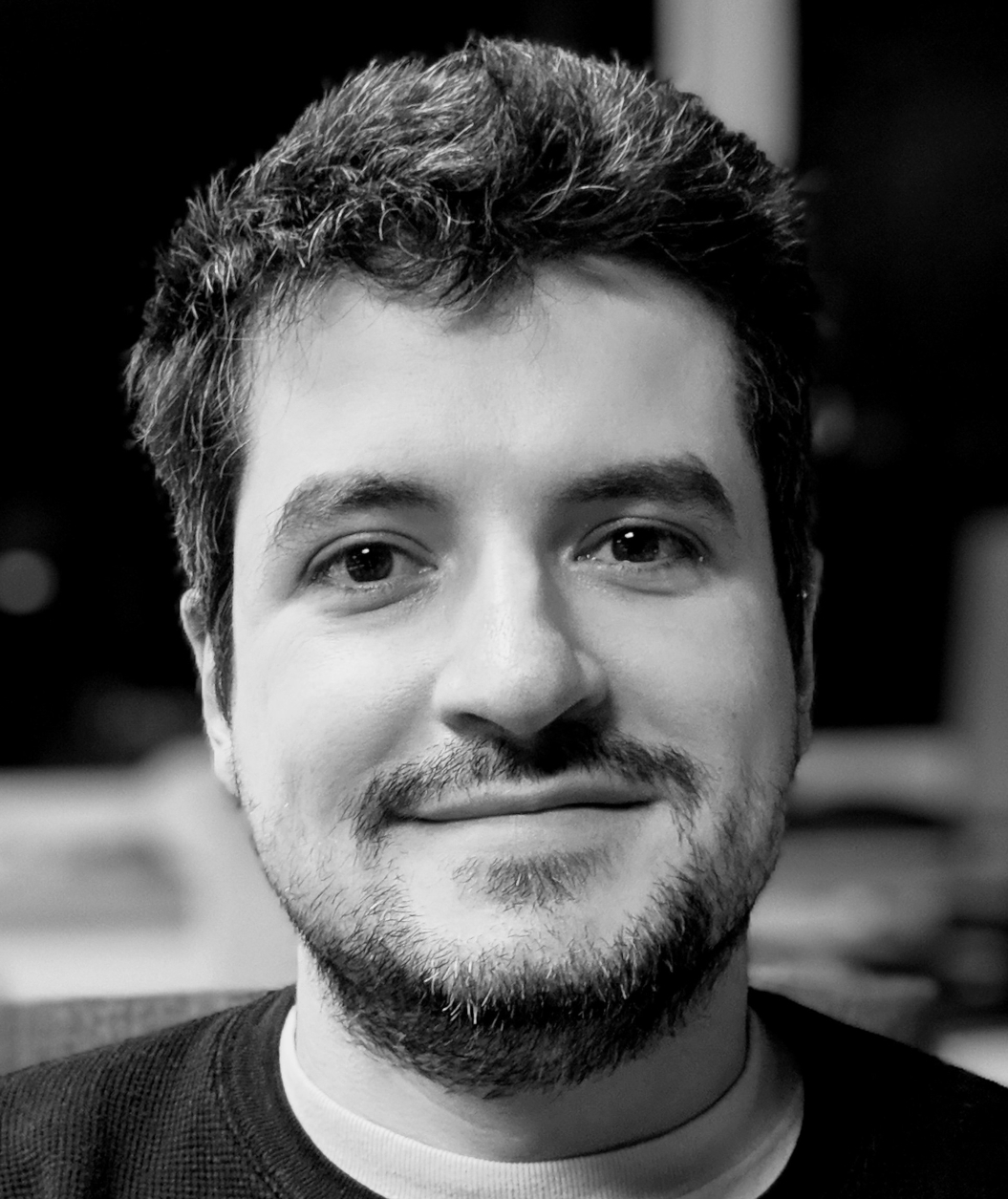}}]{Federico Tombari} is a Research Scientist and Manager at Google, where he leads an applied research team in computer vision and machine learning. Since 2014 he has been Senior Research Scientist first, currently Lecturer (PrivatDozent) at the Technical University of Munich. He has 200+ peer-reviewed publications in the field of computer vision and machine learning and their applications to robotics, autonomous driving, healthcare and augmented reality. He got his PhD in 2009 from the University of Bologna, where he was Assistant Professor from 2013 to 2016. In 2018-19 he was co-founder and managing director of a Munich-based startup on 3D perception for AR and robotics. He regularly serves as Chair and AE for international conferences and journals (ECCV18, 3DV19, ICMVA19, 3DV20, IROS20, ICRA20, RA-L among others). He was the recipient of two Google Faculty Research Awards (in 2015 and 2018), an Amazon Research Award (in 2017), 2 CVPR Outstanding Reviewer Awards (2017, 2018). His works have been awarded at conferences and workshops such as 3DIMPVT'11, MICCAI'15, ECCV-R6D'16, AE-CAI'16, ISMAR '17.
\end{IEEEbiography}

\end{document}

%% file: sections/1-intro.tex
\section{Introduction}

Video understanding is an active research field with many applications such as video recommendation, safety analysis from security cameras, human-robot interaction, etc. \cite{hutchinson2021video}. However, even though recognizing activities in trimmed videos has significantly progressed with deep learning, action segmentation, \ie, the localization and recognition of actions in untrimmed videos, has only partially benefited from it. This limitation is primarily due to the challenges and effort required to collect human-annotated large-scale data for video action segmentation, particularly frame-wise action class labels \cite{bojanowski2014weakly,stein2013combining}. Therefore, weaker forms of supervision, such as transcript supervision \cite{kuehne2014language,richard2018neuralnetwork,li2019weakly,chang2019d3tw,chang2021learning}, are commonly used to reduce the considerable effort required to obtain frame-wise action class labels for videos. In this formulation, first outlined in \cite{kuehne2014language}, methods aim to fit (or `decode') a transcript containing an ordered list of action instances of unknown length to a sequence of frames. Notably, obtaining the ground truth list requires less effort than performing frame-level annotation and could be autonomously derived from the video metadata, such as accompanying text from instructional or food preparation videos.

\begin{figure}[t]
   \begin{center}
      \includegraphics[width=\linewidth, angle=0]{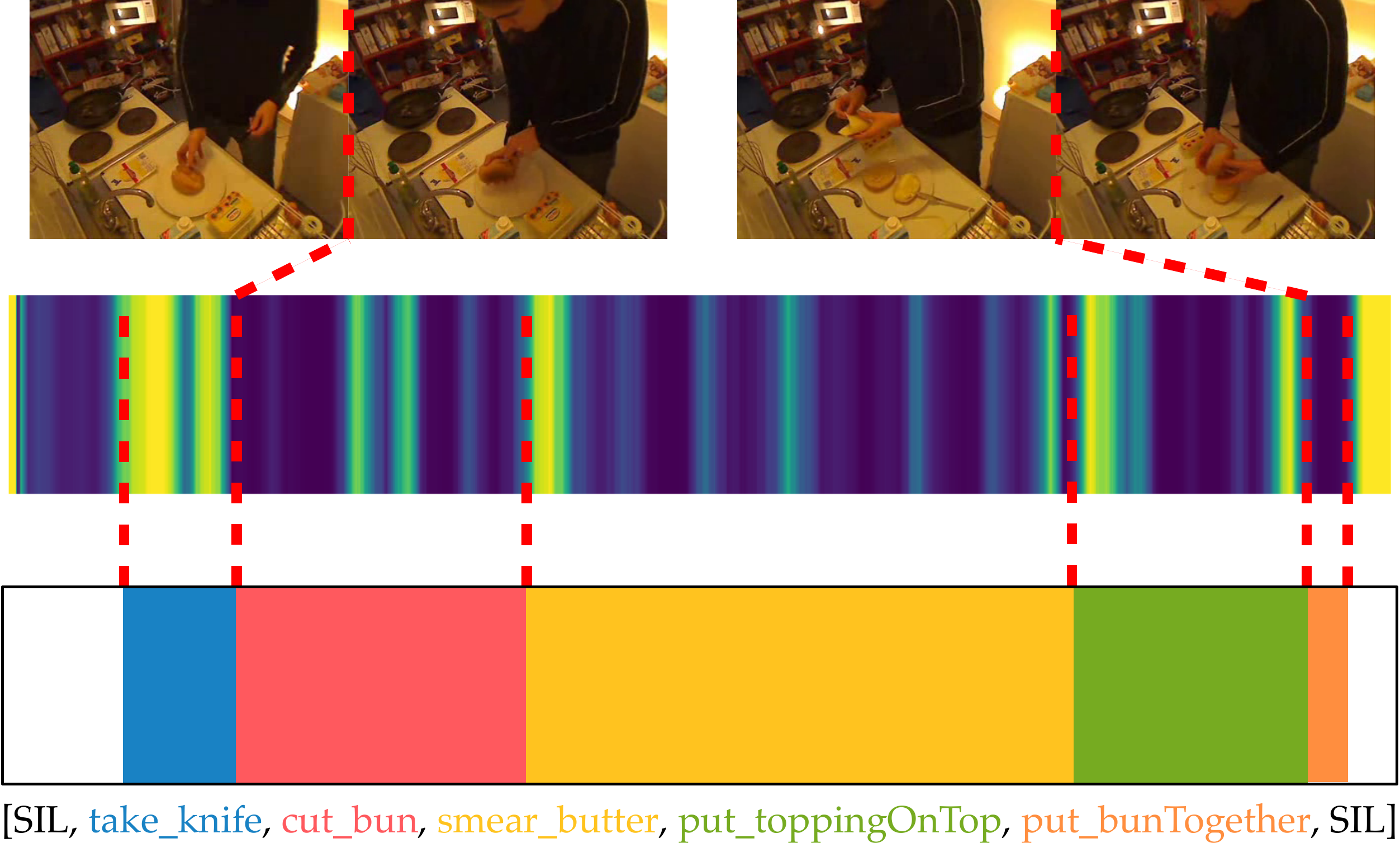}
   \end{center}
   \caption{\textbf{Modeled Attention under Transcript Supervision} - We use transformers to segment videos given only a ground-truth list of actions. The visualization of the attention mechanism illustrates how the model focuses on the starts or ends of actions comparing the attention map to the ground truth. The mechanism appears to put more emphasis on the areas around action transitions rather than the transitions themselves.
%   \cDJ{The figure shows the ground truth?} \cJo{Yes, to highlight the mechanism's focus around salient regions. It also avoids conflating the challenges with the decoding method, addressed in the decoding ambiguities figure.}
   }
   \label{fig:attention}
\end{figure}

Most of the existing methods utilize either probabilistic (Viterbi) \cite{kuehne2014language,richard2018neuralnetwork, li2019weakly} or discriminative (dynamic time warping) \cite{chang2019d3tw,chang2021learning} methods to perform the alignment of the transcript to the video frames. A temporal feature extractor is required to process the pre-extracted frame features to model temporal dependencies in these frame sequences, learning how to recognize the contiguous actions. In almost all cases, this is achieved with gated recurrent units (GRUs) \cite{cho2014learning}, a variant of recurrent neural networks (RNNs). The decoding process takes the features from the RNN, which are combined with a specific target transcript to provide the final alignment of the transcript actions. Unlike other sequence modeling tasks, transcript-supervised action segmentation introduces various challenges. Firstly, since action contiguity is assumed in the transcript, decoding methods focus on finding the optimal action lengths such that actions transition in relevant regions in the video. This consideration makes these regions, which are underrepresented in the scope of the video, significantly more critical for the saliency of a sequence modeling architecture. Secondly, when considered individually, frames lose most of their contextual meaning as actions occur over tens or hundreds of frames. Finally, the video sequences can vary significantly from hundreds to thousands of frames, meaning that modeling must be scalable. Noteworthy to say, these factors only consider the alignment of a known transcript to frames, representing only one part of the task.

Motivated by the recent trends \cite{souri2021fast}, we highlight a limitation with the existing methods surrounding the selection of unknown transcripts, which are necessary for decoding. Since the exact transcript for segmentation is not known at inference time, different methods \cite{kuehne2014language,richard2018neuralnetwork,li2019weakly,chang2019d3tw,chang2021learning} have adopted a brute-force alignment approach, where the best alignment is found for every possible training transcript. This method is not only computationally intensive \cite{souri2021fast}, \ie~significantly slowing down inference, but also only considers alignment for the transcripts, which ignores the input features and the potential ambiguities between transcripts.

For the sequence modeling required for action alignment, we propose using transformers rather than the existing RNNs, as it appears to be better suited to address some of the challenges specific to this task. Firstly, the self-attention mechanism in the transformers is better posed to characterizing `salient' regions in sequences, where attention can discern parts of the sequence that are most relevant to the task. For example, we demonstrate in \figref{fig:attention} how our transformer model applies attention to notable points in the video sequence denoting the starting and ending points of the actions. Secondly, unlike RNNs, transformers operate on multiple sequence elements simultaneously rather than relying on a propagated hidden state, which is better poised to recognize actions that occur across ranges of frames.

While the quadratic time and space complexity of the original transformer models \cite{vaswani2017attention} present challenges when scaling transformers to long sequences, variants of transformers with convolutional input operations have been successfully tailored for videos in a fully-supervised segmentation context \cite{yi2021asformer}, delivering state-of-art results. However, in recent efforts to make transformers more scalable,  \cite{child2019generating,kitaev2020reformer,wang2020linformer,katharopoulos2020transformers} address the quadratic properties of the transformer attention through modifications to reduce the space and time complexity, enabling them to scale better with more inputs. One formulation of a linear-attention transformer model \cite{katharopoulos2020transformers} has shown how exploiting associative operations inside the attention mechanism can be used to create attention that scales linearly, facilitating the use of transformers for a wider range of sequence modeling tasks normally reserved for RNNs.

With the advent of these linear-attention transformers, we demonstrate how our method, \methodname, can benefit the usage of various pre-extracted video features for weakly-supervised action segmentation. Furthermore, to adapt the transformers for the specific requirements of transcript-supervised temporal extraction, we use a sliding window of the input sequence, which controls the amount of local context encoded, similar to how different feature scales are captured in other vision transformers \cite{liu2021swin}. This technique also encourages a certain degree of monotonicity to maintain temporal correspondence between the input and output sequences, which has previously been achieved with regularization in other transformer sequence modeling tasks \cite{rios2021biasing}. Finally, the ability of the transformer to parallelize these window inputs and the lack of window-level iteration additionally reduces the time complexity of the temporal encoding compared with RNNs. We demonstrate how these models can match or exceed the action alignment performance of existing probabilistic methods through our experiments. In addition to investigating transformer-specific design considerations, such as the inclusion of positional encoding, we also test various input formulations and input feature types to gauge their compatibility with the probabilistic modeling. Through this testing, we can more thoroughly consider their role in a weakly-supervised pipeline for temporal feature extraction, which existing works have often overlooked. 

Additionally, we address transcript selection for action segmentation. Since the transcript decoder fits a specific transcript to the video, decoding requires knowing a transcript before generating an output segmentation. While ground-truth transcripts are available at training time or for action alignment, transcripts must be sourced elsewhere for inference, generally from the set of transcripts observed during training. Thus, an appropriate transcript has to be selected. Most existing approaches do not explicitly focus on this selection task, instead opting for a brute-force approach testing every possible training transcript \cite{richard2018neuralnetwork,li2019weakly,chang2019d3tw,chang2021learning}. While this method can select suitable transcripts, it is computationally intensive \cite{souri2021fast} and only considers an alignment metric to choose the transcript, which obfuscates the original feature input. Recently, a novel technique \cite{souri2021fast} has been proposed to generate the transcripts but requires a specialized transcript decoding method and cannot extend existing established approaches. To address these issues, we propose a transcript embedding approach that directly uses the video features to deliver a subset of transcripts into the decoding technique for inference, significantly reducing the number of transcripts that need to be aligned. Our method embeds the video features into an embedding space trained to capture similarities present in the videos' transcripts, achieved by taking a discriminative approach, similar to recent works on this task \cite{chang2019d3tw,chang2021learning}. A contrastive loss \cite{hadsell2006dimensionality} imbues the similarities of transcripts into the embedding, clustering together videos with similar transcripts. For inference, our method samples the embedding of a video with an unknown transcript to obtain similar videos with known transcripts, similar to image retrieval tasks \cite{deng2011hierarchical}. Furthermore, the embedding directly uses video features for transcript selection while also enabling the existing transcript scoring method to select the final transcript from a small set of candidates. We demonstrate how our extension provides significant inference speed improvements and can choose better transcripts given a diverse training set and suitable features. The selection of better transcripts ultimately improves the action segmentation performance when combined with transformer-driven action segmentation.

To summarize our contributions as follows:
\begin{itemize}
   \item We propose and demonstrate the application of linear-attention transformers to pre-extracted features on the weakly-supervised action segmentation task.
   We test various transformer ablations to understand their usage and configuration more thoroughly as temporal feature extractors on video features. We demonstrate how the use of transformers on this task reduces the number of trainable parameters required and can also meet or exceed the performance of existing probabilistic RNN-based methods.
   \item We introduce a transcript selection approach based on embeddings of video input features, using the trained embedding to select subsets of transcripts for inference. We outline how the usage of this technique can speed up existing brute-force selection approaches by multiple orders of magnitude and even increase segmentation accuracy on a suitable number of embeddings. Finally, we test our methods individually and jointly on a range of pre-extracted video features and datasets and examine their implications for this task.
\end{itemize}

In the next sections, we review the existing body of work in \secref{sec:related}, we present the proposed transformer and transcript embedding methods in \secref{sec:method}, we illustrate experiments and ablation studies in \secref{sec:experiment}, finally we draw  conclusions in \secref{sec:conclusion}.

%% file: sections/2-related.tex
\section{Related Work}
\label{sec:related}
Our methods represent the intersection of the action segmentation task under transcript supervision using transformers for video understanding. Additionally, sentence embedding and retrieval techniques motivate our transcript embedding extension.

\vspace{10pt}
\noindent
\textbf{Transcript-Supervised Action Segmentation}:
Early works in this direction used Hidden Markov Models (HMMs) to align the transcript with the video frames for transcript supervision \cite{kuehne2014language,richard2017weakly}. However, the HMMs were later replaced with Recurrent Neural Networks (RNNs) \cite{richard2018neuralnetwork} while retaining the probabilistic Viterbi decoding algorithm and class-wise priors. Due to instabilities during training on pseudo-ground truth labels, a graph-based energy loss function was proposed in a follow-up work \cite{li2019weakly} in order to facilitate training of multiple possible alignments simultaneously. Other works have additionally considered using numerical optimization \cite{bojanowski2014weakly}, alternative sequence loss representations such as Connectionist Temporal Classification (CTC) \cite{huang2016connectionist}, or train iteratively in a fully-supervised fashion on a pseudo-ground truth \cite{ding2018weakly}, but overlook the modeling of the transcript itself, which causes instability and notably impacts their performance. Recently, there has been a shift toward discriminative alignment methods with differentiable forms of Dynamic Time Warping (DTW) \cite{chang2019d3tw} or modeling transcript 'prototypes' in an embedding space \cite{chang2021learning} which have been able to outperform Viterbi-based methods slightly. Despite this, the lack of focus on the inference process has motivated recent works to replace existing approaches to address how transcripts can be generated \cite{souri2021fast} or more quickly aligned at inference-time \cite{souri2021fifa}. Conversely, we instead focus on addressing these areas with extensions to existing methods, delivering improved inference-time performance and benefits with existing pipelines and weakly-supervised paradigms.

\vspace{5pt}
\noindent
\textbf{Transformers for Video Understanding}:
Recently, transformers are starting to become more commonplace across various image and video-based tasks \cite{dosovitskiy2020image,liu2021swin,khan2021transformers,arnab2021vivit,zhang2021vidtr,neimark2021video,zheng2021rethinking}. On more conventional fully-supervised video tasks, such as action recognition \cite{arnab2021vivit,zhang2021vidtr,neimark2021video}, transformers have demonstrated improvements over existing convectional approaches when working with raw video inputs. Despite the specific formulation of the input and usage of existing features remaining an open topic \cite{neimark2021video}, transformers that use the more scalable linear attention \cite{wang2020linformer,katharopoulos2020transformers} are compatible with action anticipation tasks \cite{girdhar2021anticipative}. For action segmentation, the use of pre-extracted video features has been shown to offer state-of-art performance for fully-supervised action segmentation with I3D features \cite{yi2021asformer}. However, this has yet to be explored for other pre-extracted video features, such as the IDT features often employed in transcript-supervised action segmentation. From an unsupervised perspective, transformers have also been explored as a means to obtain latent representations of objects in videos \cite{kabra2021simone}, again indicating their ability for spatial and temporal feature extraction. Despite this, weakly-supervised applications of transformers are only starting to be investigated for video applications. Specifically, they are still grounded in tasks related to natural language processing \cite{khan2021transformers}, with approaches considering specific tasks such as video captioning \cite{tan2021logan}. Given the developments with transformers, our approach considers their application to existing transcript-supervised action segmentation and the advantages it can deliver over more conventional temporal sequence feature extractors.

\vspace{5pt}
\noindent
\textbf{Sentence Embedding/Retrieval}: 
Existing works have considered the task of defining metrics for quantifying, ranking and retrieving similar images \cite{chechik2010large}, images with specific properties \cite{siddiquie2011image} or hierarchical similarities \cite{deng2011hierarchical,barz2019hierarchy}. A deep metric learning approach \cite{wang2014learning,coskun2018human} is often taken, using contrastive or discriminative methods \cite{hadsell2006dimensionality} to train a similarity metric into the embedding. For application into videos, action segmentation methods have often considered embedding at the frame level to retrieve frames with similar actions in unsupervised \cite{sener2018unsupervised,kukleva2019unsupervised} methods, or recently the aforementioned DTW-based transcript \emph{prototypes} \cite{chang2021learning}. While our transcript embedding is motivated by these existing methods, we notably consider a single embedding of complete videos, applying similarity measures at the transcript level rather than the frame level.

%% file: sections/3-method.tex
\section{Methods}
\label{sec:method}
% \begin{figure}[t]
%   \begin{center}
%       \includegraphics[width=0.7\linewidth, angle=0]{figures/pipeline.pdf}
%   \end{center}
%   \caption{\textbf{Transcript Supervised Pipeline} - We use a probabilistic approach to decode transcripts to features encoded through the sliding window and transformer. Additionally, our transcript embedding can select transcripts from the same input features. 
%   \cDJ{This was never used in the text -- delete the figure?}
%   }\label{fig:pipeline}
% \end{figure}

Action segmentation involves the frame-wise classification of actions from a set of possible actions $\mathcal{A}$. Under transcript supervision, only an ordered list of $N$ actions $a_{1..N} \in \mathcal{A}$ is available during training. 
The objective then is to find the correspondence between the frames and the ordered list. We denote the length $\hat{l}_{n}$ as the number of frames associated to each action $a_{n} \in \mathcal{A}$.

Since the raw input frames are too unwieldy for this task, it is customary to use pre-extracted frame features taken using handcrafted or neural network methods with no additional task-specific pre-training \cite{kuehne2014language,richard2018neuralnetwork,chang2019d3tw,li2019weakly,chang2021learning}. These features offer lower dimensionalities but notably retain the temporal dimensions of the original video. Thus, for a video of length $T$, there is a set of pre-extracted features $x_{1..T}$.

\subsection{Architecture}
\begin{figure*}[t]
    \centering
   \includegraphics[width=0.8\linewidth, angle=0]{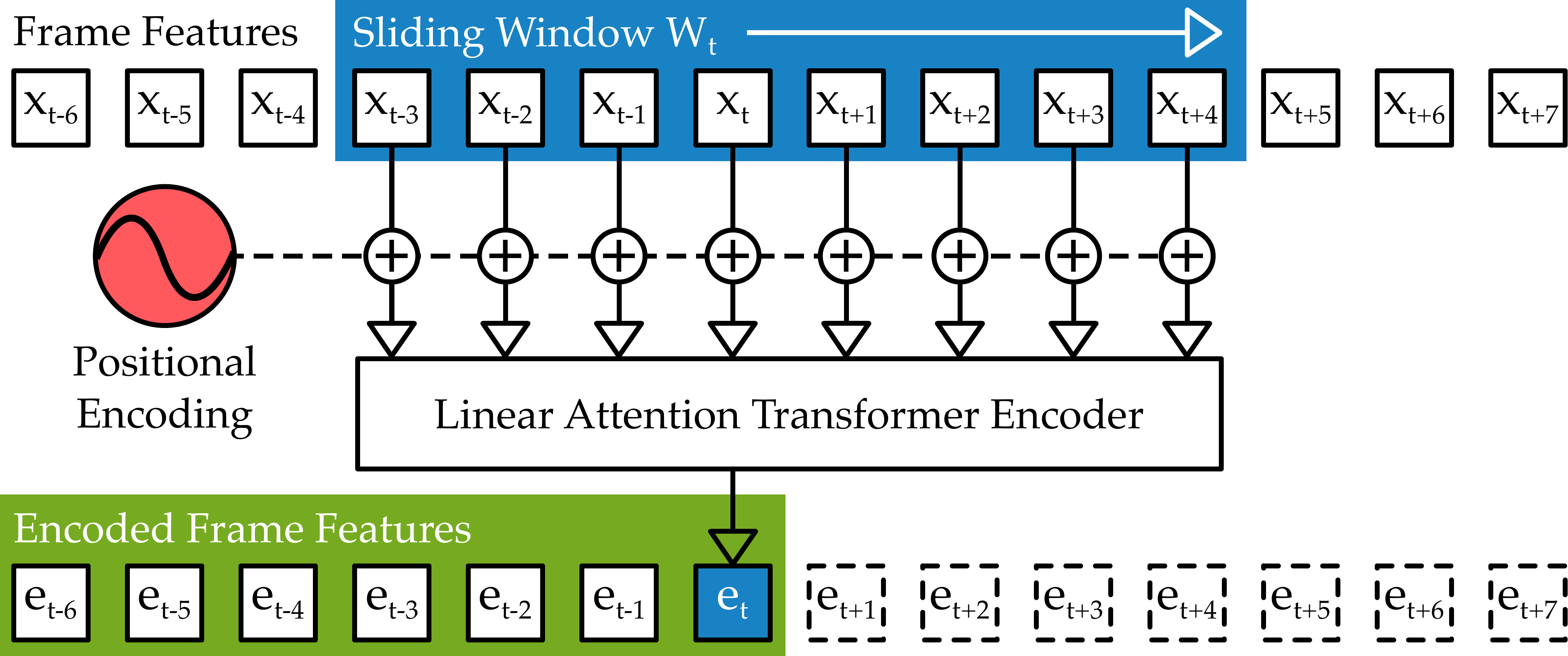}

   \caption{\textbf{Temporal Window Transformer Encoder} - We apply the linear attention transformer on successive windows of frame features to imbue local temporal context into the features. The use of a sliding window resembles a convolutional operation, but the transformer facilitates context-aware attention of window elements. We apply sinusoidal positional encoding to enable the transformer to gauge relative window positions.
%   \cDJ{This figure is too large. I think this should fit half the page.}
%   \cHu{We check small version, it was not realy look nice. I think double column is fine.}
   }
   \label{fig:transformer}
\end{figure*}

We build a linear-attention transformer \cite{katharopoulos2020transformers}, which models temporal dependencies among the frame features. \figref{fig:transformer} illustrates how our architecture outputs a sequence of features $e_{1..T}$ that considers the relationships within a constrained temporal scope, controlled by using a sliding window on the input. The encoded features are linearly projected to the number of classes where softmax is applied to produce posterior probabilities $p(a_n|e_t)$.

Our architecture, called \methodname, consists of three components: input windowing, feature encoding transformer, and probabilistic decoding.

\vspace{10pt}
\noindent
\textbf{Windowing:} 
%
% \cDJ{Maybe windowing can be separated from the architecture, i.e. a separate subsection before/after transformers?  I would suggest after since you talk about transformers.}\cJo{Not sure how easy this would be, since it very much is a precursor to the transformer in the architecture, and crops up in Fig 2 and the transformer description}
%
We utilize a sliding window on the pre-extracted features $x_{0..T}$ as the input for the subsequent transformer encoder, as shown in \figref{fig:transformer}. While this appears counterproductive when considering the transformer's ability to model long term sequences, it is necessitated by the decoding stage to facilitate the convergence of the probabilistic models, as we demonstrate in our ablation experiments.

Rather than modeling the entire sequence, the window models actions from temporally adjacent frames, providing local context to be used with probabilistic decoding techniques, thus avoiding over-fitting. With the formulation
\begin{equation}
    W_S(x, i) = \{x_j \;:\; \lfloor i - S/2 \rfloor \leq j < \lfloor i + S/2 \rfloor \} ~,
   \label{eq:window}
\end{equation}
the window $W$, defined at position $i$, has the size of $S$ where the sequence extremities are padded with zeros.
In the context of transformers, such windowing can be viewed as a temporal scoping of the attention mechanism. Furthermore, it constrains the transformer toward monotonicity, as desired in other transformer applications \cite{rios2021biasing}, which additionally ensures that the outputted features are only computed from a local temporal context.

For parallelization purposes, a single video sequence $x_{0..T}$ can be considered a stack of windows representing all the window positions with a stride of one frame. This stack is represented as
\begin{equation}
    X_W(x_{0..T}) = \text{stack}(\{W_S(x, i) \;:\; 0 \leq i < T\}),
   \label{eq:stack}
\end{equation}
which has a shape of $T \times S \times D$ for a sequence length of $T$, window size of $S$ and feature dimensionality of $D$.
These stacks are the input of the subsequent transformer encoder, represented through batches of windows. 
While the sliding window formulation enables practical input of the localized frame contexts, it presents an additional challenge in processing a long sequence of frames. As the window has a one frame stride,  there is a complete window input for each frame, which increases the overall input size by a factor of the window size.

Each window $W_t$ is used to produce a single output at the encoder, representing a single encoded frame $e_t$. Thus, processing the batches of input windows $X_W$ yields the encoded frame features, $e_{1..T}$. 
The need to parallelize motivates the use of a linear attention transformer model for this task, which can work in a time complexity of $O(T)$ for processing the windowed inputs, compared with $O(S \times T)$ for window size $S$ in an RNN. 

\vspace{10pt}
\noindent
\textbf{Linear Attention Transformer Encoder:} 
We utilize a linear transformer encoder $\xi$ to process batches of windows obtained from the input video sequence for our temporal feature extractor.
Since we are taking a single output from the input window, we only need to use the encoder portion of the architecture, as is utilized for classification tasks \cite{dosovitskiy2020image}. 

Similar to \cite{vaswani2017attention,khan2021transformers}, we use positional encoding to imbue the input features with window-relative temporal context. We use the sinusoidal positional encoding with the the positional encoding matrix $P \in \mathbb{R} ^ {S \times D}$ where
\begin{equation}
   P_{s, d} =
   \begin{cases}
      \sin(s / (10000 ^ {d/D}))           & 
      \text{if $d$ is even} \\
      \cos(s / (10000 ^ {d/D}))  & \text{otherwise},
   \end{cases}
   \label{eq:pe}
\end{equation}
that is computed at window index $s$ and feature dimension $d$, given the feature dimensionality $D$.
We can then write the transformer operation that takes the windows to yield frame-wise encodings as
\begin{equation}
    E = \xi(X_W + P),
   \label{eq:transformer}
\end{equation}
%
% \cDJ{$E(X)$ or $E(X_W)$?}\cJo{E is simply a variable to denote the output of the transformer}
where $\xi(\cdot)$ represents the encoder and $E$ represents the encoded features $e_{1..T}$. Note that the dimensionality of $E$ is ${T \times D}$, which is the same shape as the input features $x_{1..T}$.

We build the encoder with a series of multi-head self-attention layers~\cite{vaswani2017attention}, attached to a feed-forward network.
% Each encoder layer consists of a multi-head self-attention stage. The output of which is passed through a feed forward network. Skip connection and normalization succeed each both of these stages to permit the training of encoders consisting of multiple layers. 
%
The self-attention mechanism follows the comparison of the query, values and keys structure.
% The attention mechanism in the conventional transformer computes an output at a sequence position given a weighted average of all the input features, also determined from the input sequence. 
%
We obtain them by linearly projecting the input sequence to the matrices $Q$, $V$ and $K$ for the query, values and keys, respectively. 
In order to find the correlation of an arbitrary feature $i$ to the rest of the features, we use linearized attention \cite{katharopoulos2020transformers}, which considers this similarity operation by applying a learned kernel $\phi$ to both the queries and keys, as represented with
\begin{equation}
    A_i = \frac{\sum^{N}_{j=1}\phi(Q_i)^\top \phi(K_j) \cdot V_j}{\sum^{N}_{j=1}\phi(Q_i)^\top \phi(K_j)}.
   \label{eq:linearattention}
\end{equation}
%
% The linear projections of queries $Q$, values $V$ and keys $K$ are obtained from the input sequence, and a similarity function is used between the queries and keys to determine the weighting of values.
%
Considering that the queries $\phi(Q_i)^\top$ are separable from the summation of keys and values, this sum can be computed a single time and reused for a sequence, avoiding the quadratic operation of calculating the query-key similarity for each sequence element. This simplification results in the mechanism achieving both linear space and time complexity,  increasing the scalability of transformers to sequence inputs.

Since the temporal encoding for weakly-supervised segmentation is intended to provide low-level feature extraction in the temporal context, we use a single layer transformer with four self-attention heads, which is notably shallower than other applications \cite{yi2021asformer}. 
%
% \cDJ{What does thie mean? Do you mean: the transformer has one layer of self-attention with four heads?}
%
Nevertheless, it reflects the suitability of the basic attention mechanisms for the temporal feature extraction used under transcript-supervision. We examine the effects of various architectural hyper-parameter considerations in our subsequent ablation experiments. 

Consequently, an MLP head projects the encoded features into $|\mathcal{A}|$ dimensions that represent each of the possible action classes. A softmax operation for each frame provides a class posterior probability $p(a_n|e_i)$, facilitating the compatibility with the probabilistic transcript decoding technique.

\vspace{10pt}
\noindent
\textbf{Transcript Decoding:} 
We implement a decoder optimizing the alignment between the frame-wise class posterior probability with the ordered list of actions in the transcript by recursively maximizing the posterior of the constrained Viterbi algorithm \cite{kuehne2014language}. 
%
% Decoding the list of actions in the transcript $a_{1..N}$ to the class posteriors from the encoded frame features $e_{1..T}$ is achieved by recursively maximizing the posterior of the constrained Viterbi algorithm \cite{kuehne2014language}. 
%
Since the frame features and transcript actions are known, the optimization solves for the length of each transcript element $l_{1..N}$. %, which represent the final decoding. 
The recursive formulation used to decode sequence elements up to $n$ is outlined with
\begin{equation}
    \begin{split}
            p(a_{1:n}, \hat{l}_{1:n} | e_{1:t}) = \max_{t',\; t'<t} \bigg\{ p(a_{1:n-1}, \hat{l}_{1:n-1} | e_{1:t'}) \\
            \left(\prod^t_{s=t'} p(e_s | a_{n(s)})\right) p(l_n=t-t'|a_n)\bigg\}.  
    \end{split}
   \label{eq:viterbi}
\end{equation}
This formulation relies on two external probabilistic inputs, the frame-wise action likelihoods $p(e_i|a_n)$ and class-wise length model $p(l_n|a_n)$. In line with the existing methods \cite{richard2018neuralnetwork}, we define the length model with a class-wise Poisson distribution, updating $\lambda$ after each decoding. Since the decoder works with frame-wise class likelihoods, the posterior probabilities are converted using a prior $p(a_n)$ that is also updated after each encoding, as described by
\begin{equation}
    p(e_i | a_n) = \frac{p(a_n|e_i)}{p(a_n)}.
   \label{eq:bayes}
\end{equation} 
Finally, the optimal transcript element lengths $\hat{l}$ provide not only the final frame-level segmentations but are also used to construct segmentation graphs such as those used in the graphical loss function imposed while training the network.

\subsection{Loss Function}

\begin{figure}[t]
   \begin{center}
      \includegraphics[width=\linewidth, angle=0]{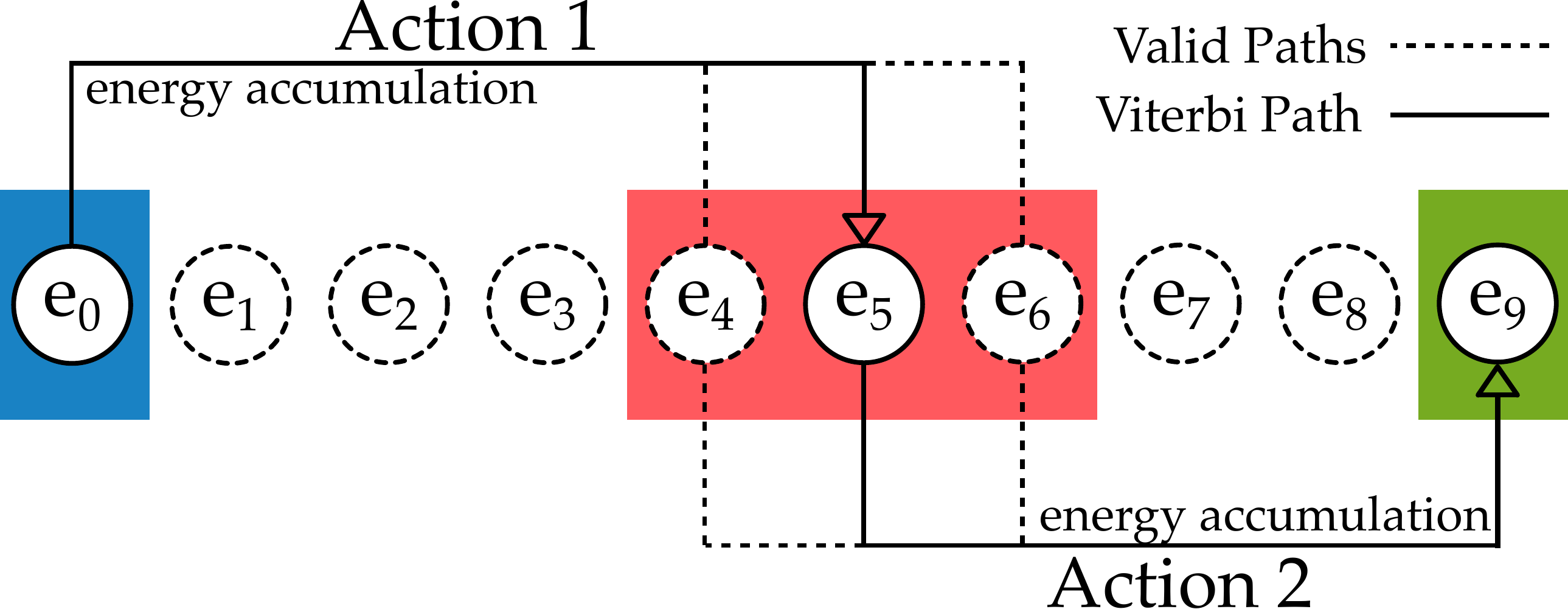}
   \end{center}
   \caption{\textbf{Forward Loss Computation} - Forward loss computation for a video transcript pair with two actions. Energy is accumulated along graph edges at action boundaries in the segmentation. Alternative segmentations in windows at these boundaries are also computed, enabling the forward loss to represent a set of valid segmentations.}
   \label{fig:cdfl}
\end{figure}

To train the transformer, we use the Constrained Discriminative Forward Loss (CDFL) \cite{li2019weakly}, originally used for training in its eponymous work. 
As shown in \figref{fig:cdfl}, this energy-based loss function operates on a graphical representation of the video segmentation, where the video frames (vertices) are segmented through transcript elements (directed edges) connecting the starting and ending points of the action. The edges represent energy that accumulates based on the frame-wise class labels. Since the loss operates directly on the network outputs, the energy minimization can be back-propagated into the network. This configuration implies training the encoder to minimize the loss function.

Repeating this process for all the edges along the segmentation path provides an energy value for the total segmentation path. An initial path is created from the current transcript decoding with edges corresponding to transcript items. While this alone represents the decoder's current state, the method additionally includes other segmentation paths to capture variations of alternative transcript decodings. Specifically, additional paths are created with neighboring frames in fixed windows at each action transition, providing slight variations to the segmentation. The set of valid paths $\mathcal{P}_V$ represents various valid segmentations with which the network is trained to identify. The path energies are approximated recursively with a $\text{logadd}(\cdot)$ operation, avoiding the exponential branching of computing each path individually. The energy from the valid paths constitutes the forward component of the loss.

While the forward loss is the primary loss component used to train the temporal feature encoding, CDFL also incorporates a constrained discriminative component. Whereas the forward loss encourages frames to be representative of valid segmentation paths, the discriminative component discourages the frames from representing invalid segmentation paths. Due to the significantly greater number of possible invalid segmentation paths $\mathcal{P}_I$, a constrained set $\mathcal{P}_{I^c}$ prevents the invalid paths from overwhelming the valid paths. Specifically, only invalid paths where the energy is lower for an incorrect class, compared to a correct class, are used. The forward and discriminative components are combined in a contrastive fashion to produce the final loss function, \begin{equation}
    \mathcal{L}_{\text{CDFL}} = \text{logadd}(\mathcal{P}_V) - \text{logadd}(\mathcal{P}_{I^c}).
   \label{eq:cdfl}
\end{equation}

The transcript segmentation pipeline is trained online due to the combination of the neural network and the probabilistic components. Rather than using batches, a single video-transcript pair is used for each iteration. Network back-propagation and reparameterization of length models and priors occur following each decoding.

\subsection{Transcript Inference}
The training and decoding processes are coupled directly to a given transcript, which is required to perform decoding. This requirement is not an issue during training, where the ground-truth transcript is known. However, it presents a challenge when the transcript is unknown, \ie~during inference. The inherent need for a transcript creates an additional task surrounding transcript selection at this stage.

Existing approaches \cite{richard2018neuralnetwork,li2019weakly,chang2019d3tw,chang2021learning} locate an \emph{optimal transcript} $\hat{t}$ from the set $\mathcal{T}$ such that 
\begin{equation}
   \hat{t} = \arg\min_{t \in \mathcal{T}} \Psi(t).
   \label{eq:selection}
\end{equation}
finds the minimum alignment cost (from Viterbi or DTW) $\Psi$.
%
% This method requires the alignment to be performed for every possible transcript, which significantly increases the runtime performance in the case of these polynomial-time algorithms. Furthermore, the selection is a by-product of the alignment and is not directly driven by the input features, which can result in unsuitable transcripts that significantly impact the segmentation accuracy on certain videos. The importance of transcript selection is quantified in the gap between known-transcript action alignment and selected transcript action segmentation scores, which indicates how transcript selection increases the overhead in performance. Additionally, it indicates the potential improvement range for action segmentation given an improved transcript selection method.
%
The importance of transcript selection is quantified in the gap between known-transcript action alignment and selected transcript action segmentation scores, which indicates how transcript selection increases the overhead in performance. 
This method requires the alignment to be performed for every possible transcript, which significantly increases runtime performance in the case of these polynomial-time algorithms. 

Furthermore, the selection is a by-product of the alignment and is not directly driven by the input features, potentially resulting in unsuitable transcripts that significantly impact the segmentation accuracy on certain videos. Thus, it is possible to improve segmentation performance immediately by improving transcript selection.

\vspace{10pt}
\noindent
\textbf{Transcript Similarity:} Realistically, a suitable transcript would be as close as possible to a video's ground truth. To quantify what makes a transcript \emph{suitable}, we outline a metric for transcript similarity \cite{souri2020evaluating}. As a list of actions is synonymous with a textual string, it is possible to apply similar methods for similarity measures. We demonstrate a normalized form of edit distance extended to compare any two arbitrary sequences, $\mathcal{X}$ and $\mathcal{Y}$, in 
\begin{equation}
   \text{sim}(\mathcal{X}, \mathcal{Y}) = 2 \frac{\text{mat}(\mathcal{X}, \mathcal{Y})}{|\mathcal{X}| + |\mathcal{Y}|}
   \label{eq:sim}
\end{equation}
where $\text{mat}(\cdot, \cdot)$ represents the size of the longest common substring, which can be calculated with a variant of the Levenshtein distance \cite{levenshtein1966binary}. Due to the normalization, a score of zero would indicate transcripts with no shared actions, whereas one would indicate identical transcripts.

With this metric, the goal of a transcript selection technique is to maximize the similarity of proposed transcripts relative to ground truth. As a means to train the network 

\vspace{10pt}
\noindent
\textbf{Transcript Embedding:} Since transcripts belong to videos as a whole, our approach obtains a single embedding at the video level, given the frame-wise input features. Thus, a suitable architecture is needed to encode the features into the embedding.

The tendency for RNNs to excessively distill information at the frame level motivated our approach to use transformers for temporal feature encoding. However, for the context of transcript embedding, we can instead utilize this limitation of RNNs to capture a more simplistic global representation of frame features, which are more useful for inter-video rather than intra-video usage.

Specifically, we use a bidirectional GRU to process the frame features and reduce the frame-wise outputs to 64 dimensions as other embedding approaches \cite{chang2021learning}. Since videos contain thousands of consecutive frames, there are often redundant representations of actions between adjacent frames, so we then downsample temporally to ensure a small but diverse representation of actions. Specifically, for frame features $X$ at rate $r$ and a random offset $\eta$ where $0 \le \eta < r$, the downsampling selects frames as shown by
\begin{equation}
   D(X) = \{x_i \in X : \forall i = c\eta, \; (c \in \mathbb{Z}) \land (i < |X|)\}.
   \label{eq:downsample}
\end{equation}

The addition of the random offset $\eta$ ensures that we can utilize all video features while also serving as a regularization technique. With this reduced formulation, we obtain a single vector by taking the mean of the sequence of embeddings, producing a single embedding point for the whole video. This method is similar to other sequence-condensing methods, such as those for sentences of words in natural language processing \cite{lin2017structured}.

\begin{figure}[t]
   \begin{center}
      \includegraphics[width=\linewidth, angle=0]{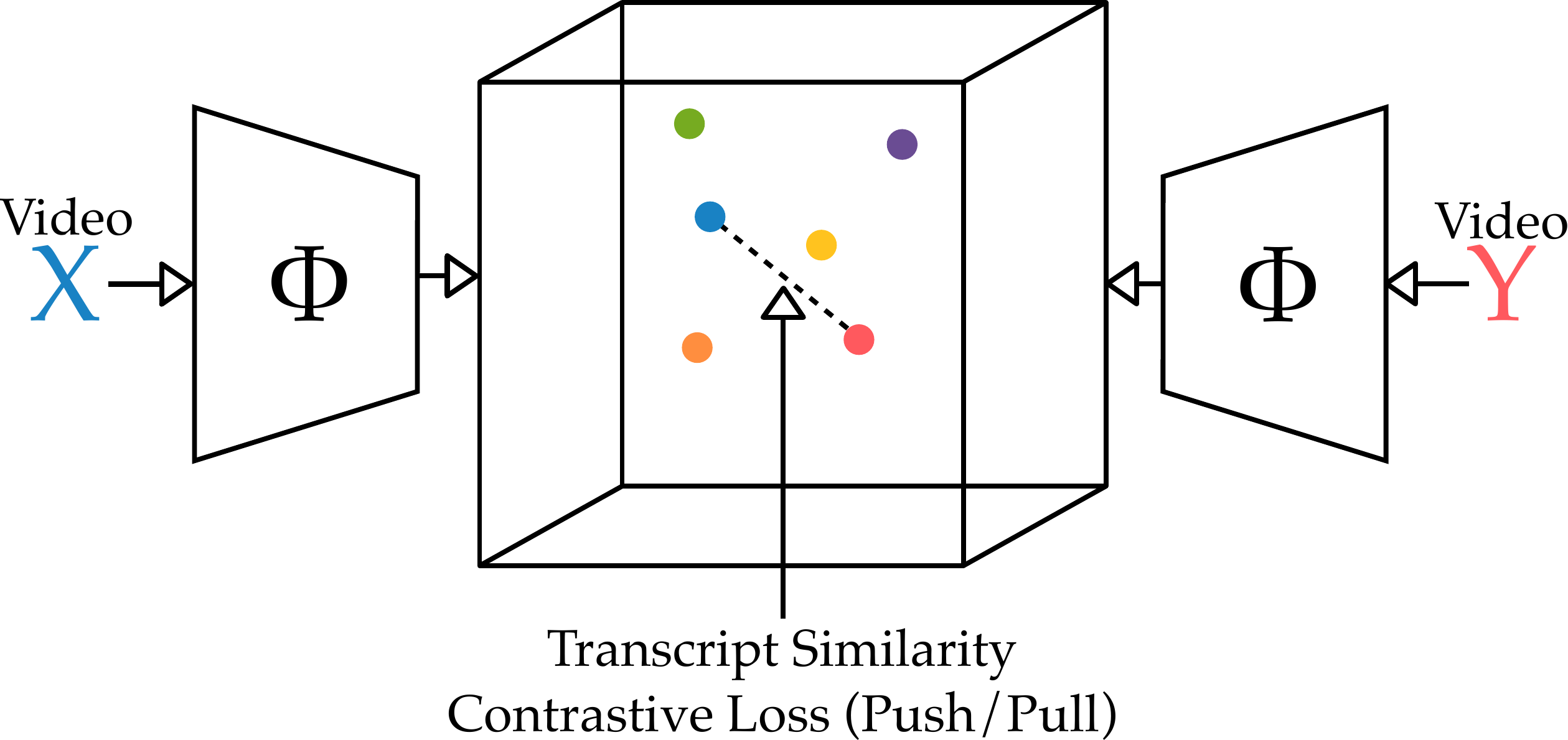}
   \end{center}
   \caption{\textbf{Video-Transcript Similarity Embedding} - We embed features from the whole video into an embedding space that is trained contrastively to reflect transcript similarity. Points in the embedding are pushed together or pulled apart based on a threshold of their transcript similarity.}
   \label{fig:embedding}
\end{figure}

For an RNN $\Theta$, we can obtain a 64 dimension embedding vector for the whole video $X$ with the process in
\begin{equation}
   \Phi(X) = \frac{\sum_{x \in D(\Theta(X))} x}{|D(X)|}.
   \label{eq:embed}
\end{equation}
Through this technique, we compare any two videos in the embedding, potentially of varying length and with different transcripts. With this embedding space, the goal is to train the network so that videos with similar transcripts are closer together than videos with different transcripts, which are further apart.

We then apply a contrastive loss function \cite{hadsell2006dimensionality} for this task to train the network toward this representation. Specifically, we define a pair-wise loss function,
\begin{multline}
   \mathcal{L}_{c}(\Phi(X_1), \Phi(X_2)) \\
   = \begin{cases}
      d(\Phi(X_1), \Phi(X_2))              & \text{if } \text{sim}(\mathcal{X}_1, \mathcal{X}_2) \geq \sigma \\
      \max\{0, m-d(\Phi(X_1), \Phi(X_2))\} & \text{otherwise,}
   \end{cases}
   \label{eq:embedloss}
\end{multline}
which can quantify the embedding similarity of two videos relative to a threshold $\sigma$ on their transcript similarity. As the function to quantify the embedding similarity $d$, we selected the cosine distance due to its constrained range and pedigree on sequence embedding \cite{xing2015normalized}. Depending on the transcript similarity, the loss encourages clustering in the embedding within this margin, as shown in \figref{fig:embedding}.

In its independent form, we train the transcript embedding in a batch-wise offline configuration. Specifically, we randomly pair videos ($X_1$, $X_2$) in batches of size 64 without replacement, which yields the loss function
\begin{equation}
   \mathcal{L}_{P} = \frac{\sum_{(X_1, X_2) \in P} \mathcal{L}_{c}(\Phi(X_1), \Phi(X_2))}{|P|}
   \label{eq:batchloss}
\end{equation}
for the set of pairs $P$.

Once trained, it is possible to exploit the embedding to retrieve suitable transcripts at inference time. Rather than having to perform alignment for every single training transcript, our method can use the trained transcript embedding to obtain a smaller subset $\mathcal{N} \subset \mathcal{T}$. This set ideally contains more relevant transcripts for alignment scoring. We create $\mathcal{N}$ by sampling nearest neighbors from the embedding up to a neighborhood size of $k$, such that $|\mathcal{N}| = k$. This process is expressed by 
\begin{equation}
   \mathcal{N} = \text{T} \left(\arg\min_{V \in \mathcal{V}} d(\Phi(I), \Phi(V))\right),
   \label{eq:inference}
\end{equation}
on the inference video $I$ for the set of training videos $\mathcal{V}$ for ground-truth transcript mapping $T$. In practice, we statically retain the embeddings of the training videos after training.

Once the reduced subset $\mathcal{N}$ has been obtained, it can substitute the original transcript set $\mathcal{T}$ in \eqref{eq:selection} for the final transcript and alignment to be determined. Thus, in addition to reducing the number of alignments required, the selection of the transcript subset is directly driven by the input of the inference video frame features. As previously mentioned, this can be done independently of the existing action segmentation pipeline and hence can apply to any previous segmentation methods that use the brute-force approach \cite{kuehne2014language,richard2018neuralnetwork,li2019weakly,chang2019d3tw,chang2021learning}.

%% file: sections/4-experiment.tex
\section{Experiments}
\label{sec:experiment}

We perform a variety of experiments with ablations to investigate various factors for the application of transformers to weakly supervised action segmentation, alignment, and transcript selection embedding. 
%
% \cDJ{I would consider removing/revising the next parts of this paragraph + next paragraph:}
% %
% We start by investigating the usage of the transformers with linear attention applied to the conventional pre-extracted IDT features in a raw fully-supervised case, demonstrating their applicability to these features. We then apply our proposed transformer formulation under weak-supervision, comparing with the existing probabilistic weakly-supervised models based on RNNs. 
% We also investigate various parts of our configuration, including transformer hyper-parmeters, window configuration, positional encoding and input features, that affect performance on this task.

% For the proposed transcript embedding, we evaluate the transcript similarity of the embedding to the existing brute-force method. Finally, we combine the weakly-supervised transformer and embedding and examine their effects on a standard datasets used for weakly-supervised action segmentation.

% \subsection{Setup}

\vspace{10pt}
\noindent
\textbf{Datasets:} 
In line with the existing works \cite{richard2018neuralnetwork,li2019weakly,chang2019d3tw,chang2021learning,souri2021fast}, we also evaluate on the Breakfast \cite{kuehne2014language}, Hollywood Extended \cite{bojanowski2014weakly}, and 50 Salads \cite{stein2013combining} datasets to compare against the state of the art. However, for ablation studies, we perform all our evaluations on the Breakfast \cite{kuehne2014language} dataset since it contains a large set of 1712 food preparation instances with 48 different action classes in videos with an average duration of 2 minutes. In contrast, the Hollywood Extended dataset is taken from human interactions in movies, with 937 videos, 16 action classes and shorter videos, averaging 30 seconds; while, the 50 Salads dataset offers a notably smaller set of only 50 food preparation instances with 17 actions but has longer videos, averaging 5 minutes. Notably, we follow the earlier works for the training and evaluation regimes. 

\vspace{10pt}
\noindent
\textbf{Metrics:} To evaluate the performance, we measure the frame accuracy (F-acc.), the ratio of correctly classified frames, as it is the primary metric and most widely used for evaluation in existing works. We additionally report the frame alignment accuracy ignoring the background frames for our final results. This formulation corrects the over-representation of the background classes in the dataset, \eg~Hollywood Extended. Outside of segmentation metrics, we also use the mean of the transcript similarity measure (TSim) in \eqref{eq:sim} as a way to measure the transcript selection performance.

\vspace{10pt}
\noindent
\textbf{Features:} 
% Unlike most existing weakly-supervised methods \cite{richard2018neuralnetwork,li2019weakly,chang2019d3tw,chang2021learning}, we experiment with a set of different pre-extracted frame feature types. 
%
We evaluate the effects of utilizing different pre-extracted features on the frames. This includes the 64 dimension PCA-projected Fisher vectors of improved dense trajectory features (IDT), which are the most established for transcript-supervision. However, the handcrafted nature of these features, along with the computational cost of their extraction, has driven the use of alternative network-derived features \cite{souri2021fast}, particularly under full-supervision \cite{li2019weakly}. We also experiment with features extracted from a pre-trained I3D \cite{carreira2017quo} network, including isolation of the specific RGB and optical flow features to show the usability of motion compared with visual features. Finally, we also test with features from a newer SlowFast \cite{feichtenhofer2019slowfast} network, which only makes use of RGB input. Where possible, we use existing extractions of the features to ensure equivalent comparison with previous methods.

\vspace{10pt}
\noindent
\textbf{\methodname{} Parameters:} 
For alignment, we use a single layer transformer with four self-attention heads, each with 64 dimensions and a feed-forward dropout of 0.5. A sliding window of 32 frames was used as an input with linear projections down to 64 dimensions for I3D and SlowFast features. A sinusoidal positional encoding was applied by window position. Transcript embedding was achieved with a 512 dimension bidirectional GRU using a transcript similarity threshold of 0.8 for Hollywood Extended and 0.5 for all other datasets. We trained both the networks with an Adam optimizer with a learning rate of 0.001.

\subsection{Raw Fully-Supervised Action Segmentation}

Before integrating the linear-attention transformer with the transcript-supervised pipeline, which requires application-specific modifications, this evaluation directly compares the transformers' performance to an existing RNN under raw full supervision, without any refinement, post-processing, or regularization stages. 
% The approaches that focus specifically on the fully-supervised case \cite{li2020ms,yi2021asformer} make use of such mechanisms to further refine the output such that the segments are encouraged to become contiguous and spurious regions are filtered out. \cDJ{Something wrong with the previous sentence.} Thus, we cannot make the comparisons in this particular experiment to these existing methods. 
Using the frame-wise class labels on the Breakfast dataset, we train the networks using a simple frame classification loss and evaluate the raw output. The transformer is configured with four layers, each with four attention heads at a dimensionality of 256 while the RNN uses gated recurrent units with a hidden dimensionality of 1024.

\input{tables/naive}

\tabref{tab:naive} describes the benefits of the linear transformer over the similarly configured RNN. 
We note that the transformer can outperform the RNN with fewer parameters while providing inference speedup for individual videos. 
Although the results indicate the compatibility of linear transformers with the frame features, we cannot directly apply this particular configuration to the transcript-supervised case without the sliding window input. We examine the reasoning behind this in our subsequent window ablation experiment presented in \tabref{tab:window}.

\subsection{Transcript-Supervised Action Segmentation}

% \cDJ{I have a feeling that this entire paragraph is  detached. I would consider removing this paragraph.}
% In addition to the use of the sliding window input for weak supervision, we also adjust the transformer by reducing it to a single layer and 64 dimension heads while retaining the four attention heads. This reduction is consistent with that of RNNs in \cite{richard2018neuralnetwork,li2019weakly}, which use 64 hidden dimensions. This parameter reduction is important to ensure all windows produced for a single video can be processed in parallel, which is necessary for this task.

With our \methodname{} architecture, we train the network under transcript supervision, in addition to implementations of the existing NN-Viterbi and CDFL methods. \tabref{tab:baseline} compares the alignment and segmentation performance of the three methods and indicates a frame alignment performance increase compared to the RNN-based methods. 
This improvement is exemplified in \figref{fig:alignment}, which examines the alignment output of various techniques. Consistent with the attention ability shown in \figref{fig:attention}, \figref{fig:badalignment} additionally shows how the attention map is able to broadly identify regions representing the start and end of actions. However, even with this ability, failure cases still exist where quick action transitions can create some ambiguity that the decoding must resolve.

\input{tables/alignment}

\input{tables/baseline}

While our method has demonstrated suitably high transcript alignment accuracy, we performed a series of ablation experiments to determine each architectural consideration's relative performance impact.

\vspace{10pt}
\noindent
\textbf{Effect of Window Size:} We investigated a variety of window sizes to determine their impact on performance. The windowing determines the scope that attention in the transformer can deviate from a pure monotonic focus, thus controlling the receptive field of each embedded output frame. We document the impact of various window sizes in \tabref{tab:window}. 
\input{tables/window}
The use of larger windows appears to produce better alignment results with diminishing returns, so long as there are sufficient computational resources to process the large windows. However, testing without the window, similar to the input in the fully-supervised case, indicates one of the limitations specific to probabilistic decoding. As the network is trained online with the probabilistic models, a balance needs to be maintained during training. Specifically, as visualized in \figref{fig:accplot}, in cases where we trained the transformer on the whole sequence, there was a tendency for it to fixate on iterative states of the probabilistic optimization, which would feed back into the probabilistic stage, causing degenerate alignment solutions.

\begin{figure}[t]
  \centering
  \begin{tikzpicture}
    \begin{axis}[
        axis x line=center,
        axis y line=center,
        yticklabel style={/pgf/number format/fixed},
        xticklabel style={/pgf/number format/fixed},
        xlabel={Iteration},
        ylabel={Alignment Accuracy},
        x label style={at={(axis description cs:0.5,-0.1)},anchor=north},
        y label style={at={(axis description cs:-0.08,.5)},rotate=90,anchor=south},
        legend pos=north east,
        every axis plot/.append style={ultra thick},
        ymax=1,
        ymin=0,
        width = \linewidth,
        height = 0.7\linewidth]
      ]

      \addplot[themeblue] table[header=false, x index = {0}, y index = {1}]{data/trainacc.csv};
      \addlegendentry{w/ Sliding Window}
      \addplot[themered] table[header=false, x index = {0}, y index = {2}]{data/trainacc.csv};
      \addlegendentry{w/o Sliding Window}

    \end{axis}
  \end{tikzpicture}
  \caption{\textbf{Training Alignment Accuracy} - The sliding window is essential to ensure the transformer does not fixate across larger parts of the sequence. Testing without the window shows how the alignment accuracy degrades during training as the transformer drives the probabilistic models into local minima.}\label{fig:accplot}
\end{figure}

\vspace{10pt}
\noindent
\textbf{Transformer Architecture:}
In addition to the windowing stage, our method involves structuring the transformer for the weakly-supervised case. As with the window, we examine the effect of transformer hyper-parameters on the alignment accuracy. Tables~\ref{tab:heads}, \ref{tab:headnumber}, \ref{tab:layernumber}, and \ref{tab:dropout} compare various head dimensions, number of heads, layers, and dropout configurations respectively on the ultimate alignment performance.

\input{tables/configuration}
While the application of transformers to the fully-supervised case found benefits of utilizing multiple-layer transformers, the smaller windowed temporal context modeled in this formulation tends to be served through a single encoder adequately. Additionally, increasing the dimensionality and number of heads did not yield measurable performance improvements.

\vspace{10pt}
\noindent
\textbf{Positional Encoding Configuration:} Since the formulation of positional encoding is important in imbuing transformers with the ability to identify order, we consider the application of various configurations of positional encoding. Specifically, \tabref{tab:pe} compares the alignment accuracy with two common types of absolute positional encoding and applications to different parts of the input. The combined video and window positional encoding alternates between the video and the window along the feature dimensions.

\input{tables/pe}

We find that the positional encoding is necessary for this transformer application. It provides the most benefit when applied to indicate the absolute window positions rather than the absolute video positions. Furthermore, a pre-computed sinusoidal positional encoding delivers better alignment results on the continuous video features than an encoding that is learned from window position. This behaviour could be attributed to the use of the sliding window, which may prevent a suitable embedding from being learned in the constrained context of the window.

\input{tables/features2}

\input{tables/transcript}

\input{tables/combined}

\vspace{10pt}
\noindent
\textbf{Input Features:} While the transformer appears to work suitably with the IDT features, we test the architecture with the easier-to-extract network-derived features. As outlined, we test with I3D features, including RGB and optical flow variants and SlowFast features, summarizing the results in \tabref{tab:features}. We apply a linear projection to reduce the dimensionality of the network features to facilitate the usage of the transformer in the same configuration as with the IDT features. This approach is similar to linear operations applied before vision transformers \cite{dosovitskiy2020image}.

Despite clear accuracy benefits demonstrated when using network features under full supervision \cite{li2020ms}, we observe that IDT features still deliver the best performance under transcript supervision. However, compared to the reported runs of the CDFL method with I3D features \cite{souri2020evaluating}, we note that the transformers can still train on the network-derived features.
Outside of comparing handcrafted and network-derived features, the focus of the features, be it visual or motion components, would also appear to have a significant impact on overall performance. Specifically, the use of features geared toward recognising specific motion, rather than the visual components, would seem to benefit the alignment of actions in sequences. Furthermore, the additional incorporation of visual features in the combined I3D architecture slightly degrades the alignment and segmentation ability. While the SlowFast architecture attempts to bypass the need for optical flow by modelling different temporal resolutions, it would also appear that it cannot capture the required motion information that would appear to be necessary for this weakly-supervised task.

\subsection{Transcript Embedding}
To evaluate the transcript embedding selection independently from the segmentation and decoding techniques, we compare the mean similarity of the transcripts taken from the embedding to those selected through the brute-force scoring approach \cite{kuehne2014language,richard2018neuralnetwork,li2019weakly,chang2019d3tw,chang2021learning}. Additionally, we compare the embeddings generated using various input feature types we investigated before. In this experiment, detailed in \tabref{tab:transcript}, we only take the nearest neighbor from the embedding.

In contrast to the observations from the segmentation task, the network-derived features, particularly I3D features, provide a significant benefit to transcript embedding over IDT features on the Breakfast dataset. Similarly, there appears to be less reliance on motion features for accurate transcript selection, with the use of both RGB and optical flow components in I3D delivering measurable improvements over sole optical flow. Therefore, having dedicated architecture for inferring transcripts with suitable input features can significantly improve inference-time over an alignment-driven approach. Furthermore, compared with the brute-force alignment scoring, the speed of transcript selection can be increased by multiple orders of magnitude.

\subsection{Combined Segmentation with Selection}
To demonstrate how we can combine our transcript embedding technique with transcript-supervised action segmentation, we finally experiment with the augmentation of the inference. Specifically, in \tabref{tab:combined}, we compare the segmentation performance of our weakly-supervised transformer using both the existing brute-force selection method and various neighborhoods of embedded transcripts. Additionally, we compare various characteristics of existing and more established transcript-supervised methods, using the complete set of three datasets to characterize the performance under different conditions. 
Starting with alignment, we see how our approach can improve the alignment accuracy across all datasets, compared to the existing CDFL probabilistic model. As our approach does not use a discriminative alignment technique, there is likely less tolerance to variability of intra-class action lengths, due to the explicit probabilistic modeling, which is why discriminative techniques can achieve better alignment scores. 

For segmentation performed with brute-force alignment scoring, we observe how the alignment scores translate somewhat proportionally into segmentation scores, again delivering performance between probabilistic and discriminative approaches. Since our approach uses probabilistic decoding, the brute-force method inference time is similar to other methods.

The application of our embedding for transcript selection can deliver significant reductions in inference time across all datasets, particularly when smaller neighborhood sizes $k$ are used. For the Breakfast dataset, we see that a single candidate from the embedding can measurably increase the overall accuracy of segmentation. Furthermore, increasing the neighborhood size to around 50 transcripts delivers notably higher segmentation accuracy, implying that combining alignment and embedding selection techniques can provide better results. This trend is also observed for the Hollywood dataset, where a neighborhood size of 50 can deliver better segmentations despite some performance degradation using a single embedding transcript. However, transcript embedding does appear to be sensitive to specific properties of datasets.

For the 50 Salads dataset, we observe that significantly fewer videos and less consistent transcripts impair the use of transcript embedding. Thus, we observe segmentation accuracy reductions with smaller neighborhood sizes, potentially indicating a case where alignment-based selection is better suited. Additionally, we note that embeddings with IDT features, rather than I3D features, were slightly more performant for Hollywood Extended. This behaviour, along with the lesser impact of transcript embedding than Breakfast, is likely the result of the shorter transcripts in this dataset. Furthermore, the over-representation of the background class in the Hollywood Extended transcripts appeared detrimental to the embedding technique. Thus, the applicability of transcript embedding is highly dependent on a dataset having a suitably large and diverse set of transcripts to facilitate contrastive modeling.

Ultimately, the combined experiments highlight alignment improvements delivered through the use of our transformer architecture and the importance of the transcript selection component for the segmentation task.

%% file: tables/naive.tex
\begin{table}[t]
   \centering
   \begin{tabular}{l|c|c|c}
    \toprule
      \multicolumn{1}{c}{Architecture} & \multicolumn{1}{c}{Inf. Speedup} & \multicolumn{1}{c}{Param.} & \multicolumn{1}{c}{F-Acc.} \\
      \midrule
      GRU (1024-dim) & 1$\times$ & 3.3M & 49.15\\
      LSTM (1024-dim) & 0.74$\times$ & 4.5M & 43.38\\
      Bidirectional GRU (512-dim) & 0.71$\times$ & 3.6M & 50.77\\
      Bidirectional LSTM (512-dim) & 0.54$\times$ & 4.8M & 45.4\\
      Lin. Attn. Transf. (4$\times$256) & 3.4$\times$ & 2.6M & \textbf{55.31}\\
      \bottomrule
   \end{tabular}
   \caption{\textbf{Raw Fully-Supervised Action Segmentation} - Comparison of RNN and linear transformer fully-supervised on the first split of the Breakfast dataset with only a frame-wise classification loss on IDT features. Windowing is not used since as there is no need to ensure compatibility with weakly-supervised decoding. Unlike established fully-supervised methods, post-processing, refinement, and regularization operations are not performed.}\label{tab:naive}
\end{table}

%% file: tables/alignment.tex
\begin{figure*}[t]
    \begin{center}
       \includegraphics[width=\linewidth, angle=0]{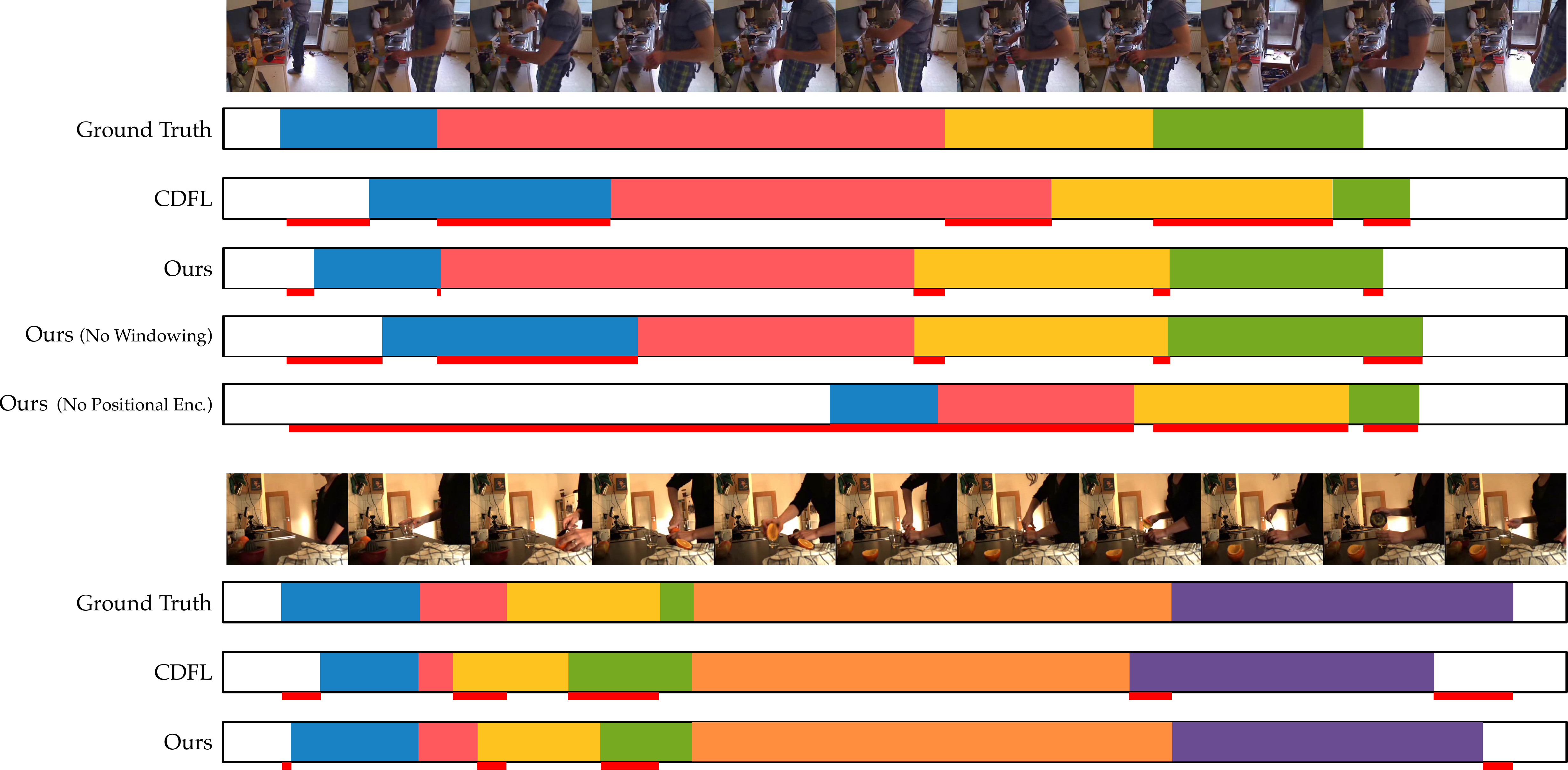}
    \end{center}
    \caption{\textbf{Alignment Comparison} - Qualitative comparison of action alignments between the ground truth, CDFL (RNN) and our transformer models. For these examples denoting cereal and juice preparation, we note that our method better aligns to action transitions than the existing CDFL method. Without windowing, we observe that action classes tend toward regions of equal length, the initialization for the probabilistic modeling. Furthermore, without positional encoding we observe that alignments degenerate completely, with most of the labels belonging to a single transcript item.}
    \label{fig:alignment}
 \end{figure*}

\begin{figure*}[t]
    \begin{center}
       \includegraphics[width=\linewidth, angle=0]{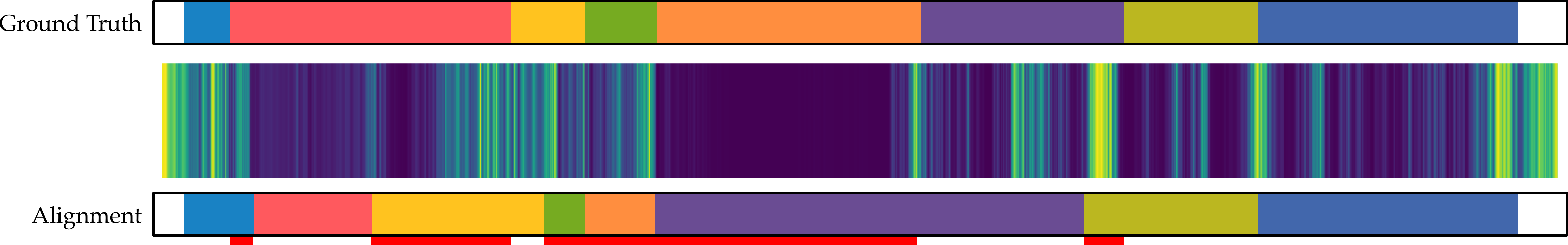}
    \end{center}
    \caption{\textbf{Decoding Ambiguities} - While the attention mechanism does appear to identify the starting and endpoints of actions, spurious responses can introduce ambiguity, which, when combined with assumptions in the probabilistic decoding, can cause misalignment of actions. This example of scrambled egg preparation, which contains many short consecutive actions, indicates how the start and end points can be confused.}
    \label{fig:badalignment}
 \end{figure*}

%% file: tables/baseline.tex
\begin{table}[b]
   \centering
   \begin{tabular}{l|c|c|c}
    \toprule
        \multicolumn{1}{c}{} & \multicolumn{1}{c}{} & \multicolumn{2}{c}{Frame Accuracy (F-Acc.)} \\  
      \cmidrule{3-4}
       \multicolumn{1}{c}{Method} & \multicolumn{1}{c}{Architecture} & \multicolumn{1}{c}{\textit{Alignment}} & \multicolumn{1}{c}{\textit{Segmentation}} \\
      \midrule
      NN-Viterbi \cite{richard2018neuralnetwork} & GRU & 55.5 & 42.5 \\
      CDFL \cite{li2019weakly}& GRU & 62.1 & \textbf{49.9} \\
      \methodname{} (Ours) & Transformer & \textbf{64.5} & 49.8 \\
      \bottomrule
   \end{tabular}
   \caption{\textbf{Transcript-Supervised Action Segmentation} - Computed action segmentation and action alignment frame accuracies on the first split of the Breakfast dataset with IDT features. Performance of existing methods has been recomputed using author-provided implementations. Use of the transformer architecture increases alignment accuracy compared to the RNN GRU architecture.}\label{tab:baseline}
\end{table}

%% file: tables/window.tex
\begin{table}[t]
   \centering
   \begin{tabular}{c|c|c}
    \toprule
      \multicolumn{1}{c}{~~Window Type~~} & \multicolumn{1}{c}{~~Size ($S$)~~} & \multicolumn{1}{c}{~~~~F-acc.~~~~} \\
      \midrule
      \multirow{4}{*}{Sliding}  & 4 & 51.8\\
       & 8 & 52.0\\
       & 16 & 64.5\\
       & 32 & \textbf{65.1}\\
      \midrule
      None & $T$ & 24.2\\
      \bottomrule
   \end{tabular}
   \caption{\textbf{Input Window Ablation} - Effect of various sliding window sizes on the Breakfast alignment accuracy. Larger window sizes increase accuracy with diminishing returns. Processing the whole sequence in a single window incorporates global context, which prevents the probabilistic models from converging to a suitable solution.}\label{tab:window}
\end{table}

%% file: tables/configuration.tex
\begin{table}[t]
   \centering
   \begin{tabular}{c|c}
    \toprule
      \multicolumn{1}{c}{Head Dimension} & \multicolumn{1}{c}{Action Alignment F-Acc.}\\
      \midrule
      32 & 57.5\\
      64 & \textbf{64.5}\\
      128 & 52.0\\
      \bottomrule
   \end{tabular}
   \caption{\textbf{Head Dimension Ablation} - Alignment under various attention head dimensions. Testing of head size 128 required reduction of window size, which impacted relative performance.}\label{tab:heads}
\end{table}

\begin{table}[t]
   \centering
   \begin{tabular}{c|c}
    \toprule
      \multicolumn{1}{c}{\# Heads} & \multicolumn{1}{c}{Action Alignment F-Acc.}\\
      \midrule
      2 & 56.2\\
      4 & \textbf{64.5}\\
      8 & 64.1\\
      \bottomrule
   \end{tabular}
   \caption{\textbf{\# Heads Ablation} - Alignment with various number of attention heads. Addition of more than four heads provided no measurable improvements for increased resource requirements.}\label{tab:headnumber}
\end{table}

% \begin{table}[t]
%   \centering
%   \begin{tabular}{c|ccc}
%     \toprule
%       \multicolumn{1}{c}{} & \multicolumn{3}{c}{Head Dimension}\\
%       \cmidrule{2-4}
%       \multicolumn{1}{c}{Number of Heads} & \multicolumn{1}{c}{32} & \multicolumn{1}{c}{64} & \multicolumn{1}{c}{128}\\
%       \midrule
%       2 & \\
%       4 & \\
%       8 & \\
%       \bottomrule
%   \end{tabular}
%   \caption{\textbf{} - }\label{tab:headablation}
% \end{table}

\begin{table}[t]
   \centering
   \begin{tabular}{c|c}
    \toprule
      \multicolumn{1}{c}{\# Layers} & \multicolumn{1}{c}{Action Alignment F-Acc.}\\
      \midrule
      1 & \textbf{64.5}\\
      2 & 62.2\\
      3 & 62.3\\
      \bottomrule
   \end{tabular}
   \caption{\textbf{\# Layers Ablation} - A single encoder layer appears to suitably model the features for the probabilistic decoding. Additional layers did not yield any further modeling improvements. }\label{tab:layernumber}
\end{table}

\begin{table}[t]
   \centering
   \begin{tabular}{c|c}
    \toprule
      \multicolumn{1}{c}{Dropout} & \multicolumn{1}{c}{Action Alignment F-Acc.}\\
      \midrule
      0.2 & \textbf{64.6}\\
      0.5 & 64.5\\
      0.8 & 64.5\\
      \bottomrule
   \end{tabular}
   \caption{\textbf{Dropout Ablation} - On this task the transformer encoder did not appear to be sensitive to different amounts of dropout.}\label{tab:dropout}
\end{table}

%% file: tables/pe.tex
\begin{table}[t]
   \centering
   \begin{tabular}{l|c|c}
    \toprule
      \multicolumn{1}{c}{Positional Encoding} & \multicolumn{1}{c}{Applied to} & \multicolumn{1}{c}{F-Acc.}\\
      \midrule
      \multirow{3}{*}{Sinusoidal \cite{vaswani2017attention}} & Video & 43.2 \\
    %   \cline{2-2}
       & Window & \textbf{64.5} \\
       & Video + Window & 64.3 \\
      \midrule
      Learned \cite{dosovitskiy2020image} & Window & 54.5 \\
      \midrule
      None & -- & 37.1 \\
      \bottomrule
   \end{tabular}
   \caption{\textbf{Positional Encoding Ablation} - Effect of various positional encoding strategies on the Breakfast alignment accuracy. Application to the window positions rather than video positions yields better results. Significant performance reduction without any positional encoding highlights its importance on this formulation.}\label{tab:pe}
\end{table}

%% file: tables/features2.tex
\begin{table}[t]
   \centering
   \begin{tabular}{c|c|c|c|c}
    \toprule
    \multicolumn{1}{c}{}    & \multicolumn{1}{c}{} & \multicolumn{1}{c}{} & \multicolumn{2}{c}{F-Acc.} \\  
    \cmidrule{4-5}
       \multicolumn{1}{c}{Type}    & \multicolumn{1}{c}{Input} & \multicolumn{1}{c}{Linear Projection} & \multicolumn{1}{c}{\textit{Alignment}} & \multicolumn{1}{c}{\textit{Segmentation}} \\
      \midrule
      IDT & Motion & None & \textbf{64.5} & 49.8\\
      \midrule
      \multirow{3}{*}{I3D} & Both & $2048 \rightarrow 64$ & 59.7 & 45.7\\
       & RGB & $1024 \rightarrow 64$ & 45.9 & 36.5\\
       & Flow & $1024 \rightarrow 64$ & \textbf{64.5} & \textbf{50.2}\\
      \midrule
      SlowFast & RGB & $2304 \rightarrow 64$ & 54.5 & 41.3\\
      \bottomrule
   \end{tabular}
   \caption{\textbf{Performance with Various Features} - Action segmentation and alignment performance using different types of pre-extracted frame features. Notably, features that focus on motion, IDT and I3D optical flow, deliver better alignment and segmentation scores.}\label{tab:features}
\end{table}

%% file: tables/transcript.tex
\begin{table}[t]
   \centering
   \begin{tabular}{c|c|c|c|c}
      \toprule
      \multicolumn{1}{c}{Method} & \multicolumn{1}{c}{Features} & \multicolumn{1}{c}{Input} & \multicolumn{1}{c}{TSim} & \multicolumn{1}{c}{Inference (sec.)} \\
      \midrule 
      Align. Scoring & IDT & Motion & 0.694 & 1884 \\
      \midrule
      \multirow{5}{*}{Embedding $k=1$} & IDT & Motion  & 0.655 & \multirow{5}{*}{0.1}\\
      \cmidrule{2-4}
      & \multirow{3}{*}{I3D} & Both &  \textbf{0.739}\\
      &  & RGB & 0.691\\
      &  & Flow & 0.716\\
      \cmidrule{2-4}
      & SlowFast & RGB & 0.714\\
      \bottomrule
   \end{tabular}
   \caption{\textbf{Transcript Embedding Performance} - Focusing on the transcript selection task with various features, we compare the suitability of transcripts taken from the embedding (single nearest neighbor) to those from the brute-force scoring method. Network derived-features appear to be more suitable for transcript selection.}\label{tab:transcript}
\end{table}

%% file: tables/combined.tex
\begin{table*}[t]
   \centering
   \begin{tabular}{c|c|c|c|c|c|c|c|c|c}
      \toprule
      \multicolumn{1}{c}{}
      &\multicolumn{1}{c}{} & \multicolumn{1}{c}{} & \multicolumn{1}{c}{} & \multicolumn{1}{c}{} & \multicolumn{1}{c}{} & \multicolumn{1}{c}{} & \multicolumn{3}{c}{Frame Accuracy (F-acc.)}     \\
      \cmidrule{8-10}
      \multicolumn{1}{c}{}
      &\multicolumn{1}{c}{Method} & \multicolumn{1}{c}{Features} & \multicolumn{1}{c}{Architecture} & \multicolumn{1}{c}{\multirow[t]{1}{*}{\shortstack{Transcript\\Selector}}} & \multicolumn{1}{c}{Alignments} & \multicolumn{1}{c}{\multirow[t]{1}{*}{\shortstack{Inference\\Time (sec.)}}} & \multicolumn{1}{c}{\textit{Alignment}} & \multicolumn{1}{c}{\textit{Align. (no BG)}} & \multicolumn{1}{c}{\textit{Segmentation}} \\
      \midrule
      \multirow{10}{*}{\rotatebox[origin=c]{90}{\centering \scriptsize{\textbf{Breakfast}}}} & 
      MuCon \cite{souri2021fast} & I3D & CNN & RNN Generator & 1 & 3 (as reported) & - & - & 49.7 \\
      \cline{2-7}
      & NN-Viterbi \cite{richard2018neuralnetwork} & \multirow{9}{*}{IDT} & \multirow{2}{*}{Prob. RNN} & \multirow{5}{*}{\shortstack{Brute-force\\Scoring}} & \multirow{5}{*}{237} & 1725 & - & - & 43.0 \\
      & CDFL \cite{li2019weakly} & & &  &  & 1884 & 63.0 & 61.4 & 50.2 \\
      \cline{4-4}
      & D3TW \cite{chang2019d3tw} & & \multirow{2}{*}{Disc. RNN} &  &  & - & 57.0 & - & 45.7 \\
      & DP-DTW \cite{chang2021learning} & &  &  &  & - & \textbf{67.7} & - & 50.8 \\
      \cline{2-2}
      \cline{4-4}
      \cline{8-9}
      & \multirow{5}{*}{Ours} & & \multirow{5}{*}{\shortstack{Prob.\\Transformer}} &  & &  1884 & \multirow{5}{*}{65.5} & \multirow{5}{*}{\textbf{62.5}} & 50.5 \\
      \cline{5-7}
      & &  & & \multirow{4}{*}{\shortstack{I3D\\Embed.}} & $k=1$ & 17 & & & 51.7 \\
      & &  & & & $k=10$ & 89 & & & 52.4 \\
      & &  & & & $k=50$ & 191 & & & \textbf{53.2} \\
      & &  & & & $k=100$ & 537 & & & 52.9 \\
      \midrule
      \multirow{8}{*}{\rotatebox[origin=c]{90}{\centering \scriptsize{\textbf{Hollywood Extended}}}}
      & CDFL \cite{li2019weakly} & \multirow{8}{*}{IDT} & Prob. RNN &   \multirow{4}{*}{\shortstack{Brute-force\\Scoring}} & \multirow{4}{*}{361} & 157 & 64.3 & 70.8 & 45.0 \\
      \cline{4-4}
      & D3TW \cite{chang2019d3tw} &  & \multirow{2}{*}{Disc. RNN} & &  & - & 59.4 & - & 33.6 \\
      & DP-DTW \cite{chang2021learning} & &  &  &  & - & \textbf{66.4} & - & \textbf{55.6} \\
      \cline{2-2}
      \cline{4-4}
      \cline{8-9}
      &\multirow{5}{*}{Ours} & & \multirow{5}{*}{\shortstack{Prob.\\Transformer}} &  &  & 157 & \multirow{5}{*}{64.8} & \multirow{5}{*}{\textbf{71.0}} & 46.2 \\
      \cline{5-7}
      & & & &\multirow{4}{*}{\shortstack{IDT\\Embed.}} & $k=1$ & 3 & & & 42.1 \\
      & &  &&&  $k=10$ & 4 & & & 45.6 \\
      & &  &&&  $k=50$ & 7 & & & 47.7 \\
      & &  &&&  $k=100$ & 10 & & & 47.2 \\
      \midrule
      \multirow{6}{*}{\rotatebox[origin=c]{90}{\centering \scriptsize{\textbf{50 Salads}}}} 
      & NN-Viterbi \cite{richard2018neuralnetwork} & \multirow{6}{*}{IDT} & \multirow{2}{*}{Prob. RNN} & \multirow{3}{*}{\shortstack{Brute-force\\Scoring}} & \multirow{3}{*}{40} & 1843 & - & - & 49.4 \\
      & CDFL \cite{li2019weakly} & & & &  & 2143 & 68.0 & 65.3 & 54.7 \\
      \cline{2-2}
      \cline{4-4}
      \cline{8-9}
      &\multirow{4}{*}{Ours} & & \multirow{4}{*}{\shortstack{Prob.\\Transformer}} &  &  & 2143 & \multirow{4}{*}{\textbf{68.2}} & \multirow{4}{*}{\textbf{65.4}} & \textbf{55.0} \\
      \cline{5-7}
      && &&\multirow{3}{*}{\shortstack{I3D\\Embed.}} & $k=1$ & 50 & & & 41.1 \\
      &&  &&& $k=10$ & 240 & & & 48.2 \\
      &&  &&& $k=25$ & 1247 & & & 51.5 \\
      \bottomrule
   \end{tabular}
   \caption{\textbf{Action Segmentation (Combined)} - We compare the segmentation and inference time performance of our methods to best reported results from existing works. Our method outperforms the existing CDFL probabilistic approach for transcript alignment. When supplemented with our transcript selection technique, we not only significantly reduce the inference duration, but provide notable segmentation accuracy improvements on both the Breakfast and Hollywood Extended datasets.}\label{tab:combined}
\end{table*}

%% file: sections/5-conclusion.tex
\section{Conclusion}
\label{sec:conclusion}
Transformer architectures, particularly linear-attention transformers, are much more suitably posed than existing sequence modeling techniques for tasks involving videos. These benefits are demonstrated through their uptake into video-related tasks and inherent ability to model salient temporal regions. However, while taking a weakly-supervised approach for a task such as action segmentation alleviates much of the effort in obtaining ground truth annotations, it increases the complexity of the associated training pipeline by imposing various constraints on architectures, including transformers. 

With the proposed architecture we have first demonstrated that it is possible to use linear attention transformers to perform transcript supervised action segmentation in the context of existing probabilistic paradigms. Under this modeling, the transformers demonstrated a greater ability to align actions to a known transcript rather than an existing RNN. 

Secondly, we proposed a video-transcript embedding to quickly identify existing videos with known transcripts, such that a suitable subset of transcripts can be applied at inference time. Through the combination of this technique with various input features, we demonstrated how such an embedding significantly reduces the inference duration. Furthermore, the method can deliver more accurate transcript selections where a suitable amount of transcripts are modeled. These improvements emphasize both the importance of transcript selection on this task and the limitations of the pre-extracted features.

Given the apparent affinity of the transformer architecture toward video modeling tasks, their more comprehensive integration could further benefit this pipeline. These integrations could include feature extraction from raw video frames, decoding frames to specific transcripts, or modeling the transcripts themselves.

%% file: main.bbl
\begin{thebibliography}{10}
\providecommand{\url}[1]{#1}
\csname url@samestyle\endcsname
\providecommand{\newblock}{\relax}
\providecommand{\bibinfo}[2]{#2}
\providecommand{\BIBentrySTDinterwordspacing}{\spaceskip=0pt\relax}
\providecommand{\BIBentryALTinterwordstretchfactor}{4}
\providecommand{\BIBentryALTinterwordspacing}{\spaceskip=\fontdimen2\font plus
\BIBentryALTinterwordstretchfactor\fontdimen3\font minus
  \fontdimen4\font\relax}
\providecommand{\BIBforeignlanguage}[2]{{%
\expandafter\ifx\csname l@#1\endcsname\relax
\typeout{** WARNING: IEEEtran.bst: No hyphenation pattern has been}%
\typeout{** loaded for the language `#1'. Using the pattern for}%
\typeout{** the default language instead.}%
\else
\language=\csname l@#1\endcsname
\fi
#2}}
\providecommand{\BIBdecl}{\relax}
\BIBdecl

\bibitem{hutchinson2021video}
M.~S. Hutchinson and V.~N. Gadepally, ``Video action understanding: A
  tutorial,'' \emph{IEEE Access}, 2021.

\bibitem{bojanowski2014weakly}
P.~Bojanowski, R.~Lajugie, F.~Bach, I.~Laptev, J.~Ponce, C.~Schmid, and
  J.~Sivic, ``Weakly supervised action labeling in videos under ordering
  constraints,'' in \emph{European Conference on Computer Vision}.\hskip 1em
  plus 0.5em minus 0.4em\relax Springer, 2014, pp. 628--643.

\bibitem{stein2013combining}
S.~Stein and S.~J. McKenna, ``Combining embedded accelerometers with computer
  vision for recognizing food preparation activities,'' in \emph{Proceedings of
  the 2013 ACM international joint conference on Pervasive and ubiquitous
  computing}, 2013, pp. 729--738.

\bibitem{kuehne2014language}
H.~Kuehne, A.~Arslan, and T.~Serre, ``The language of actions: Recovering the
  syntax and semantics of goal-directed human activities,'' in
  \emph{Proceedings of the IEEE conference on computer vision and pattern
  recognition}, 2014, pp. 780--787.

\bibitem{richard2018neuralnetwork}
A.~Richard, H.~Kuehne, A.~Iqbal, and J.~Gall, ``Neuralnetwork-viterbi: A
  framework for weakly supervised video learning,'' in \emph{Proceedings of the
  IEEE conference on Computer Vision and Pattern Recognition}, 2018, pp.
  7386--7395.

\bibitem{li2019weakly}
J.~Li, P.~Lei, and S.~Todorovic, ``Weakly supervised energy-based learning for
  action segmentation,'' in \emph{Proceedings of the IEEE/CVF International
  Conference on Computer Vision}, 2019, pp. 6243--6251.

\bibitem{chang2019d3tw}
C.-Y. Chang, D.-A. Huang, Y.~Sui, L.~Fei-Fei, and J.~C. Niebles, ``D3tw:
  Discriminative differentiable dynamic time warping for weakly supervised
  action alignment and segmentation,'' in \emph{Proceedings of the IEEE/CVF
  Conference on Computer Vision and Pattern Recognition}, 2019, pp. 3546--3555.

\bibitem{chang2021learning}
X.~Chang, F.~Tung, and G.~Mori, ``Learning discriminative prototypes with
  dynamic time warping,'' in \emph{Proceedings of the IEEE/CVF Conference on
  Computer Vision and Pattern Recognition}, 2021, pp. 8395--8404.

\bibitem{cho2014learning}
K.~Cho, B.~van Merri{\"e}nboer, C.~Gulcehre, D.~Bahdanau, F.~Bougares,
  H.~Schwenk, and Y.~Bengio, ``Learning phrase representations using rnn
  encoder--decoder for statistical machine translation,'' in \emph{Proceedings
  of the 2014 Conference on Empirical Methods in Natural Language Processing
  (EMNLP)}, 2014, pp. 1724--1734.

\bibitem{souri2021fast}
Y.~Souri, M.~Fayyaz, L.~Minciullo, G.~Francesca, and J.~Gall, ``Fast weakly
  supervised action segmentation using mutual consistency,'' \emph{IEEE
  Transactions on Pattern Analysis and Machine Intelligence}, 2021.

\bibitem{vaswani2017attention}
A.~Vaswani, N.~Shazeer, N.~Parmar, J.~Uszkoreit, L.~Jones, A.~N. Gomez,
  {\L}.~Kaiser, and I.~Polosukhin, ``Attention is all you need,'' in
  \emph{Advances in neural information processing systems}, 2017, pp.
  5998--6008.

\bibitem{yi2021asformer}
F.~Yi, H.~Wen, and T.~Jiang, ``Asformer: Transformer for action segmentation,''
  in \emph{The British Machine Vision Conference (BMVC)}, 2021.

\bibitem{child2019generating}
R.~Child, S.~Gray, A.~Radford, and I.~Sutskever, ``Generating long sequences
  with sparse transformers,'' \emph{arXiv preprint arXiv:1904.10509}, 2019.

\bibitem{kitaev2020reformer}
N.~Kitaev, L.~Kaiser, and A.~Levskaya, ``Reformer: The efficient transformer,''
  in \emph{International Conference on Learning Representations}, 2019.

\bibitem{wang2020linformer}
S.~Wang, B.~Z. Li, M.~Khabsa, H.~Fang, and H.~Ma, ``Linformer: Self-attention
  with linear complexity,'' \emph{arXiv preprint arXiv:2006.04768}, 2020.

\bibitem{katharopoulos2020transformers}
A.~Katharopoulos, A.~Vyas, N.~Pappas, and F.~Fleuret, ``Transformers are rnns:
  Fast autoregressive transformers with linear attention,'' in
  \emph{International Conference on Machine Learning}.\hskip 1em plus 0.5em
  minus 0.4em\relax PMLR, 2020, pp. 5156--5165.

\bibitem{liu2021swin}
Z.~Liu, Y.~Lin, Y.~Cao, H.~Hu, Y.~Wei, Z.~Zhang, S.~Lin, and B.~Guo, ``Swin
  transformer: Hierarchical vision transformer using shifted windows,''
  \emph{International Conference on Computer Vision (ICCV)}, 2021.

\bibitem{rios2021biasing}
A.~R. Gonzales, C.~Amrhein, N.~Aepli, and R.~Sennrich, ``On biasing transformer
  attention towards monotonicity,'' in \emph{Proceedings of the 2021 Conference
  of the North American Chapter of the Association for Computational
  Linguistics: Human Language Technologies}, 2021, pp. 4474--4488.

\bibitem{hadsell2006dimensionality}
R.~Hadsell, S.~Chopra, and Y.~LeCun, ``Dimensionality reduction by learning an
  invariant mapping,'' in \emph{2006 IEEE Computer Society Conference on
  Computer Vision and Pattern Recognition (CVPR'06)}, vol.~2.\hskip 1em plus
  0.5em minus 0.4em\relax IEEE, 2006, pp. 1735--1742.

\bibitem{deng2011hierarchical}
J.~Deng, A.~C. Berg, and L.~Fei-Fei, ``Hierarchical semantic indexing for large
  scale image retrieval,'' in \emph{CVPR 2011}.\hskip 1em plus 0.5em minus
  0.4em\relax IEEE, 2011, pp. 785--792.

\bibitem{richard2017weakly}
A.~Richard, H.~Kuehne, and J.~Gall, ``Weakly supervised action learning with
  rnn based fine-to-coarse modeling,'' in \emph{Proceedings of the IEEE
  conference on Computer Vision and Pattern Recognition}, 2017, pp. 754--763.

\bibitem{huang2016connectionist}
D.-A. Huang, L.~Fei-Fei, and J.~C. Niebles, ``Connectionist temporal modeling
  for weakly supervised action labeling,'' in \emph{European Conference on
  Computer Vision}.\hskip 1em plus 0.5em minus 0.4em\relax Springer, 2016, pp.
  137--153.

\bibitem{ding2018weakly}
L.~Ding and C.~Xu, ``Weakly-supervised action segmentation with iterative soft
  boundary assignment,'' in \emph{Proceedings of the IEEE Conference on
  Computer Vision and Pattern Recognition}, 2018, pp. 6508--6516.

\bibitem{souri2021fifa}
Y.~Souri, Y.~A. Farha, F.~Despinoy, G.~Francesca, and J.~Gall, ``Fifa: Fast
  inference approximation for action segmentation,'' \emph{arXiv preprint
  arXiv:2108.03894}, 2021.

\bibitem{dosovitskiy2020image}
A.~Dosovitskiy, L.~Beyer, A.~Kolesnikov, D.~Weissenborn, X.~Zhai,
  T.~Unterthiner, M.~Dehghani, M.~Minderer, G.~Heigold, S.~Gelly \emph{et~al.},
  ``An image is worth 16x16 words: Transformers for image recognition at
  scale,'' in \emph{International Conference on Learning Representations},
  2020.

\bibitem{khan2021transformers}
S.~Khan, M.~Naseer, M.~Hayat, S.~W. Zamir, F.~S. Khan, and M.~Shah,
  ``Transformers in vision: A survey,'' \emph{arXiv preprint arXiv:2101.01169},
  2021.

\bibitem{arnab2021vivit}
A.~Arnab, M.~Dehghani, G.~Heigold, C.~Sun, M.~Lu{\v{c}}i{\'c}, and C.~Schmid,
  ``Vivit: A video vision transformer,'' \emph{arXiv preprint
  arXiv:2103.15691}, 2021.

\bibitem{zhang2021vidtr}
Y.~Zhang, X.~Li, C.~Liu, B.~Shuai, Y.~Zhu, B.~Brattoli, H.~Chen, I.~Marsic, and
  J.~Tighe, ``Vidtr: Video transformer without convolutions,'' in
  \emph{Proceedings of the IEEE/CVF International Conference on Computer
  Vision}, 2021, pp. 13\,577--13\,587.

\bibitem{neimark2021video}
D.~Neimark, O.~Bar, M.~Zohar, and D.~Asselmann, ``Video transformer network,''
  \emph{arXiv preprint arXiv:2102.00719}, 2021.

\bibitem{zheng2021rethinking}
S.~Zheng, J.~Lu, H.~Zhao, X.~Zhu, Z.~Luo, Y.~Wang, Y.~Fu, J.~Feng, T.~Xiang,
  P.~H. Torr \emph{et~al.}, ``Rethinking semantic segmentation from a
  sequence-to-sequence perspective with transformers,'' in \emph{Proceedings of
  the IEEE/CVF Conference on Computer Vision and Pattern Recognition}, 2021,
  pp. 6881--6890.

\bibitem{girdhar2021anticipative}
R.~Girdhar and K.~Grauman, ``Anticipative video transformer,'' \emph{arXiv
  preprint arXiv:2106.02036}, 2021.

\bibitem{kabra2021simone}
R.~Kabra, D.~Zoran, G.~Erdogan, L.~Matthey, A.~Creswell, M.~Botvinick,
  A.~Lerchner, and C.~P. Burgess, ``Simone: View-invariant,
  temporally-abstracted object representations via unsupervised video
  decomposition,'' \emph{arXiv preprint arXiv:2106.03849}, 2021.

\bibitem{tan2021logan}
R.~Tan, H.~Xu, K.~Saenko, and B.~A. Plummer, ``Logan: Latent graph co-attention
  network for weakly-supervised video moment retrieval,'' in \emph{Proceedings
  of the IEEE/CVF Winter Conference on Applications of Computer Vision}, 2021,
  pp. 2083--2092.

\bibitem{chechik2010large}
G.~Chechik, V.~Sharma, U.~Shalit, and S.~Bengio, ``Large scale online learning
  of image similarity through ranking.'' \emph{Journal of Machine Learning
  Research}, vol.~11, no.~3, 2010.

\bibitem{siddiquie2011image}
B.~Siddiquie, R.~S. Feris, and L.~S. Davis, ``Image ranking and retrieval based
  on multi-attribute queries,'' in \emph{CVPR 2011}.\hskip 1em plus 0.5em minus
  0.4em\relax IEEE, 2011, pp. 801--808.

\bibitem{barz2019hierarchy}
B.~Barz and J.~Denzler, ``Hierarchy-based image embeddings for semantic image
  retrieval,'' in \emph{2019 IEEE Winter Conference on Applications of Computer
  Vision (WACV)}.\hskip 1em plus 0.5em minus 0.4em\relax IEEE, 2019, pp.
  638--647.

\bibitem{wang2014learning}
J.~Wang, Y.~Song, T.~Leung, C.~Rosenberg, J.~Wang, J.~Philbin, B.~Chen, and
  Y.~Wu, ``Learning fine-grained image similarity with deep ranking,'' in
  \emph{Proceedings of the IEEE conference on computer vision and pattern
  recognition}, 2014, pp. 1386--1393.

\bibitem{coskun2018human}
H.~Coskun, D.~J. Tan, S.~Conjeti, N.~Navab, and F.~Tombari, ``Human motion
  analysis with deep metric learning,'' in \emph{Proceedings of the European
  Conference on Computer Vision (ECCV)}, 2018, pp. 667--683.

\bibitem{sener2018unsupervised}
F.~Sener and A.~Yao, ``Unsupervised learning and segmentation of complex
  activities from video,'' in \emph{Proceedings of the IEEE Conference on
  Computer Vision and Pattern Recognition}, 2018, pp. 8368--8376.

\bibitem{kukleva2019unsupervised}
A.~Kukleva, H.~Kuehne, F.~Sener, and J.~Gall, ``Unsupervised learning of action
  classes with continuous temporal embedding,'' in \emph{Proceedings of the
  IEEE/CVF Conference on Computer Vision and Pattern Recognition}, 2019, pp.
  12\,066--12\,074.

\bibitem{souri2020evaluating}
Y.~Souri, A.~Richard, L.~Minciullo, and J.~Gall, ``On evaluating weakly
  supervised action segmentation methods,'' \emph{arXiv preprint
  arXiv:2005.09743}, 2020.

\bibitem{levenshtein1966binary}
V.~I. Levenshtein \emph{et~al.}, ``Binary codes capable of correcting
  deletions, insertions, and reversals,'' in \emph{Soviet physics doklady},
  vol.~10.\hskip 1em plus 0.5em minus 0.4em\relax Soviet Union, 1966, pp.
  707--710.

\bibitem{lin2017structured}
Z.~Lin, M.~Feng, C.~N.~d. Santos, M.~Yu, B.~Xiang, B.~Zhou, and Y.~Bengio, ``A
  structured self-attentive sentence embedding,'' \emph{arXiv preprint
  arXiv:1703.03130}, 2017.

\bibitem{xing2015normalized}
C.~Xing, D.~Wang, C.~Liu, and Y.~Lin, ``Normalized word embedding and
  orthogonal transform for bilingual word translation,'' in \emph{Proceedings
  of the 2015 Conference of the North American Chapter of the Association for
  Computational Linguistics: Human Language Technologies}, 2015, pp.
  1006--1011.

\bibitem{carreira2017quo}
J.~Carreira and A.~Zisserman, ``Quo vadis, action recognition? a new model and
  the kinetics dataset,'' in \emph{proceedings of the IEEE Conference on
  Computer Vision and Pattern Recognition}, 2017, pp. 6299--6308.

\bibitem{feichtenhofer2019slowfast}
C.~Feichtenhofer, H.~Fan, J.~Malik, and K.~He, ``Slowfast networks for video
  recognition,'' in \emph{Proceedings of the IEEE/CVF international conference
  on computer vision}, 2019, pp. 6202--6211.

\bibitem{li2020ms}
S.-J. Li, Y.~AbuFarha, Y.~Liu, M.-M. Cheng, and J.~Gall, ``Ms-tcn++:
  Multi-stage temporal convolutional network for action segmentation,''
  \emph{IEEE transactions on pattern analysis and machine intelligence}, 2020.

\end{thebibliography}
